\documentclass[11pt]{article}

\usepackage[final]{acl}
\usepackage{times}
\usepackage{latexsym}
\usepackage{listings}
\usepackage[T1]{fontenc}
\usepackage[main=english,vietnamese]{babel}
\usepackage{microtype}
\microtypesetup{protrusion=false}
\usepackage[utf8]{inputenc}
\usepackage{microtype}
\usepackage{inconsolata}
\usepackage{graphicx}
\usepackage{booktabs}
\usepackage{subcaption}
\usepackage{multirow}
\usepackage[table,xcdraw]{xcolor}
\usepackage{todonotes}
\usepackage{pifont}
\usepackage{arydshln}
\usepackage{tabularx}
\usepackage{xspace}
\newcommand{\greencheck}{\begingroup\color[HTML]{008026}\ding{52}\endgroup}
\newcommand{\redcheck}{\begingroup\textcolor{red}{\ding{55}}\endgroup}
\definecolor{highres}{HTML}{C8E6C9}    
\definecolor{medres}{HTML}{FFF9C4}     
\definecolor{lowres}{HTML}{FFCDD2}     
\definecolor{rowgray}{HTML}{F5F5F5} 
\definecolor{PosGreen}{HTML}{008026}

\usepackage{amsmath}

\usepackage[most]{tcolorbox}
\tcbset{
  promptbox/.style={
    enhanced,
    boxrule=0.6pt,
    colback=gray!3,
    colframe=gray!60,
    arc=2mm,
    left=2mm,right=2mm,top=1.2mm,bottom=1.2mm,
    fonttitle=\bfseries,
    coltitle=black
  }
}

\title{Multilingual Idioms in Sentences and Conversations Across High-, Medium-, and Low-Resource Languages}

\author{ Saeed Almheiri\thanks{\xspace\xspace Equal contribution.}\textsuperscript{1}\quad 
Bilal Elbouardi$^*$\textsuperscript{1}\quad 
Salsabila Zahirah Pranida$^*$\textsuperscript{1} \\ 
\textbf{Irina Nikishina\textsuperscript{2}}\quad 
\textbf{Ashwath Rao B\textsuperscript{3}}\quad
\textbf{Parameswari Krishnamurthy\textsuperscript{4}}\\
\textbf{Muhammad Cendekia Airlangga\textsuperscript{1}}\quad 
\textbf{Rifo Ahmad Genadi\textsuperscript{1}} \quad
\textbf{\viet{Nguyễn Phan Gia Bảo}\textsuperscript{5}}\\
\textbf{Amir Hossein Yari\textsuperscript{1}}\quad
\textbf{Hawau Olamide Toyin\textsuperscript{1}}\quad
\textbf{Nurdaulet Mukhituly\textsuperscript{1}}\\
\textbf{Mena Attia\textsuperscript{1}}\quad
\textbf{Besher Hassan\textsuperscript{1}}\quad 
\textbf{Ahmad Fathan Hidayatullah\textsuperscript{6}}\\
\textbf{Tatsuki Kuribayashi\textsuperscript{1}}\quad
\textbf{Haonan Li\textsuperscript{1}}\quad 
\textbf{Suma Bhat\textsuperscript{7}}\quad 
\textbf{Fajri Koto\textsuperscript{1}} \\
\textsuperscript{1}Mohamed bin Zayed University of Artificial Intelligence \quad
\textsuperscript{2}University of Hamburg\\
\textsuperscript{3}Manipal University \quad
\textsuperscript{4}IIIT Hyderabad \quad
\textsuperscript{5}University of Science and Technology of Hanoi\\
\textsuperscript{6}Universitas Islam Indonesia\quad
\textsuperscript{7}Princeton University \\
\texttt{\small \{saeed.y, bilal.elbouardi, salsabila.pranida\}@mbzuai.ac.ae}
} 
\begin{document}
\maketitle
\begin{abstract}
Idiomatic expressions pose a major challenge for multilingual NLP because their meanings shift between figurative and literal usage, often requiring context for accurate interpretation. Prior work has focused on high-resource languages typically evaluates isolated idiom-meaning questions, overlooking realistic discourse. We introduce MIDI, a multilingual idiom dataset spanning 3 high-, 3 medium-, and 12 low-resource languages, curated by native speakers. Unlike previous datasets, MIDI provides idioms embedded in both sentence-level and conversational contexts, capturing both literal and figurative readings. Benchmarking state-of-the-art models shows that idiom comprehension degrades in low-resource languages and that, in all resource tiers, literal interpretations are substantially harder than figurative ones. Conversational context improves performance but does not eliminate these disparities. Through controlled tests and interventions on hidden representations, we further separate memorization from reasoning, exposing core limitations of current models\footnote{The dataset can be accessed at \url{https://huggingface.co/datasets/Almheiri/MultIdiom}, and the accompanying code is available at \url{https://github.com/bitalov/multilingual_idiom}.}.
\end{abstract}


\section{Introduction}

While large language models (LLMs) demonstrate impressive capabilities, processing idioms remains a focus of ongoing research \cite{zhou-bhat-2024-non, zhou2024enhancing, kim-etal-2025-memorization} due to their semantic ambiguity between literal and figurative meanings \cite{baldwin2010multiword}. Idioms serve as a unique test bed for assessing the limits because, while their meanings can be memorized as patterns, correctly inferring them in context requires integrating nuanced cultural cues and reasoning-based inference \cite{cacciari1988comprehension, dankers-etal-2022-transformer, kovacs2016definition}.


\begin{figure}[t]
    \centering \includegraphics[width=\linewidth]{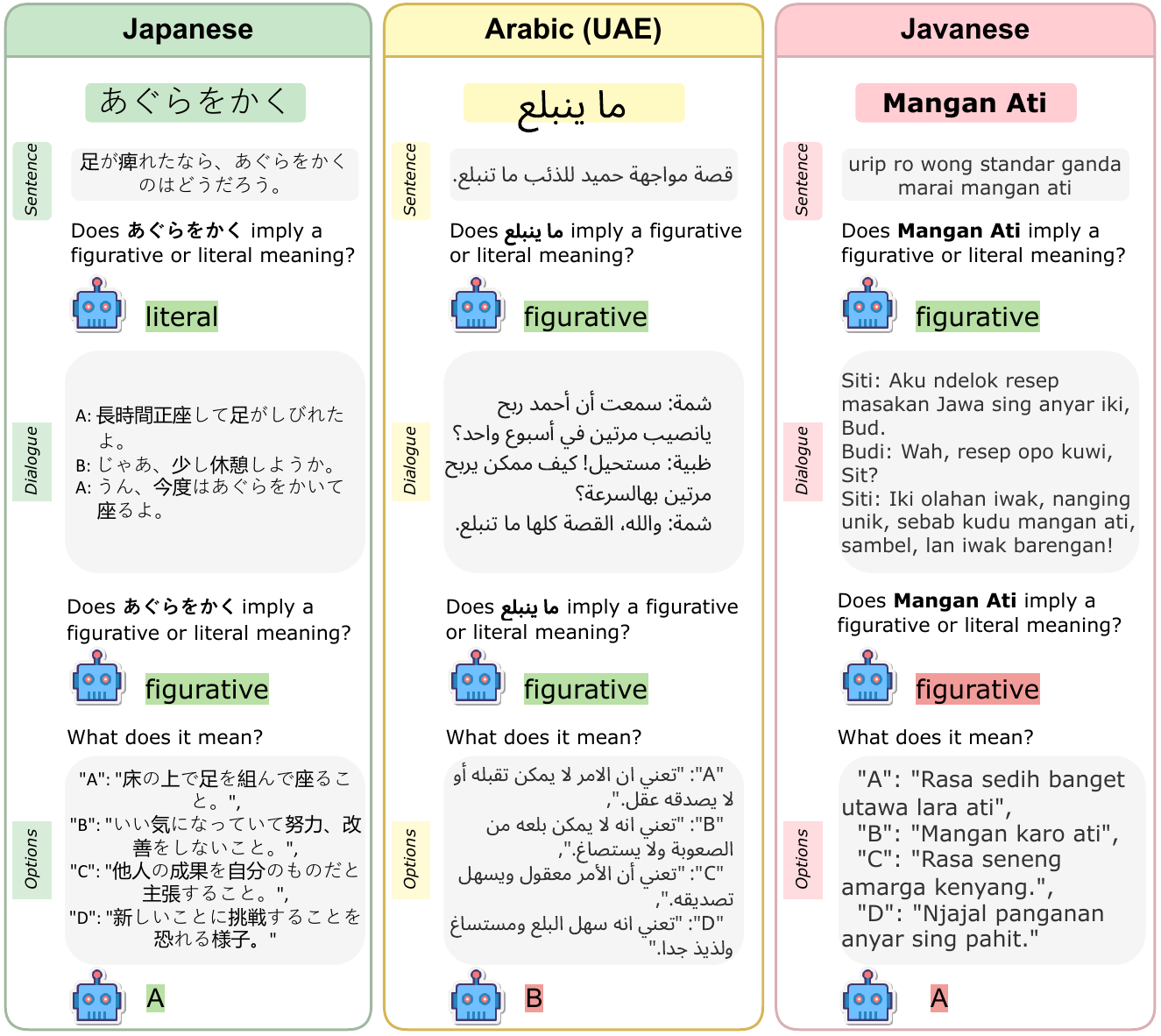}
    \caption{We compile idioms and their sentence- and dialogue-level usages from 18 languages spanning high\mbox{-,} medium-, and low-resource contexts, then evaluate LLMs with multiple-choice and binary inference tasks targeting both figurative vs. literal understanding and biased interpretations.}
    \label{fig:placeholder}
    \setlength{\abovecaptionskip}{0pt}
    \setlength{\belowcaptionskip}{0pt}
\end{figure}

Recent studies show that state-of-the-art models perform well with high-resource idioms \cite{mi-etal-2025-rolling}, but struggle in lower-resource settings where training data is sparse and deep cultural grounding is required. The current study expands this test bed to investigate the complex interplay between memorization and reasoning \cite{kim-etal-2025-memorization}. By examining how models handle these non-compositional expressions, we aim to better characterize the hybrid mechanisms that drive idiomatic understanding across diverse linguistic~landscapes.



\begin{table*}[ht!]
    \centering
    \setlength{\belowcaptionskip}{0pt}
    \resizebox{\textwidth}{!}{%
    \begin{tabular}{@{}lcccccc@{}}
    \toprule

\multirow{2}{*}{Reference} & \multirow{2}{*}{\# of Languages} & \multirow{2}{*}{MCQ} & \multicolumn{2}{c}{Context Granularity} & \multirow{2}{*}{Low Resource}  & \multirow{2}{*}{\begin{tabular}[c]{@{}c@{}}Multi-dimension \\ annotation\end{tabular}}  \\
         &  &  & Sentence & Conversation &    &  \\
        \midrule
        LIDIOMS~\cite{moussallem-etal-2018-lidioms} & 5 & \redcheck & \greencheck & \redcheck & \redcheck & \redcheck\\
        MAGPIE~\cite{haagsma-etal-2020-magpie} & 1 & \redcheck  & \greencheck & \greencheck & \redcheck & \redcheck\\
        AStitchInLanguageModels~\cite{tayyar-madabushi-etal-2021-astitchinlanguagemodels-dataset} & 2 & \redcheck & \greencheck & \greencheck & \redcheck & \redcheck\\
        ID10MS~\cite{tedeschi-etal-2022-id10m} & 10 & \redcheck & \greencheck & \redcheck & \redcheck & \redcheck \\
        MAPS \cite{liu-etal-2024-multilingual}& 6 & \greencheck & \redcheck & \greencheck & \greencheck & \redcheck \\
        Persian-MAPS \cite{khoshtab-etal-2025-comparative}& 1 & \greencheck & \redcheck & \greencheck & \greencheck & \redcheck \\
        DICE \cite{mi-etal-2025-rolling}& 1 & \redcheck & \greencheck & \redcheck & \redcheck & \redcheck \\
        MIDAS \cite{kim-etal-2025-memorization} & 6 & \greencheck & \greencheck & \redcheck & \redcheck & \redcheck \\
         \midrule
         Ours &  18  & \greencheck & {\color[HTML]{008026} \ding{52}} & {\color[HTML]{008026} \ding{52}} & {\color[HTML]{008026} \ding{52}} & {\color[HTML]{008026} \ding{52}} \\
         \bottomrule
    \end{tabular}
    }
    \caption{\textbf{MIDI vs.\ prior idiom benchmarks} across languages, context, coverage, and annotations.}
    \label{tab:datasets}
\end{table*}

Existing multilingual benchmarks are limited in their language diversity or the discourse environments where idioms naturally occur (see Table \ref{tab:datasets} for more details). Besides, current multilingual large language models (mLLMs) have not been systematically tested on their ability to reason over the idioms semantic ambiguity across diverse linguistic and cultural contexts. This leads us to pose the following research questions that we expect to answer with a view of diverse linguistic landscapes: \textit{(1) Do models interpret idioms by memorizing their meanings, or by reasoning from the existing context to disambiguate usage? (2) Do they generalize across various languages, or does idiom disambiguation remain language-specific? (3) To what extent does context---sentential or conversational---help models distinguish between figurative and non-figurative meaning?}

To address these questions, we curate a multilingual idiom dataset \textbf{MIDI} spanning \textbf{18 languages and dialects}.  Following the taxonomy established by \citet{joshi-etal-2020-state}, we categorized these languages into three tiers based on their digital availability and institutional support: 
high-resource \textit{(Chinese, Japanese, Russian)}, medium-resource (\textit{Indonesian, Vietnamese, and the UAE dialect of Arabic}) languages demonstrating moderate digital presence and are identified as "rising stars" in global NLP benchmarks \cite{costa2022no}, and 12 low-resource languages and regional dialects \textit{(Javanese, Persian, Kannada, Telugu, Tamil, Minangkabau, Sundanese, Kazakh, Yoruba, Arabic [Syrian, Egyptian, and Moroccan dialects])}, selected due to their severe underrepresentation in standard training sets and their reliance on nuanced cultural grounding for idiomatic interpretation.


MIDI contains over 100 idioms per language, selected and validated by native speakers and annotated with key psycholinguistic properties, including familiarity, literal plausibility, and idiom decomposability \cite{LibbenTitone2008}. Importantly, the idioms in the dataset are embedded in two realistic discourse settings: (1) sentence-level contexts, each crafted in both literal and figurative senses; and (2) short dialogues, where pragmatic cues, speaker intentions, and conversational flow influence interpretation. Each instance is paired with manually verified multiple-choice comprehension questions for controlled evaluation.

To further investigate memorized idiomatic knowledge vs. genuine reasoning, we provide parallel evaluation splits of ``memorization'' and ``contextual reasoning,'' inspired by recent methodology that separates definitional recall from usage-based inference \cite{kim-etal-2025-memorization}. By probing models on figurative/literal classification both with and without prior access to idiom definitions, we evaluate current systems' abilities to interpret idioms beyond rote recall.

In summary, our contributions are as follows: (i) we introduce \textbf{MIDI}, a broad testbed spanning 18 typologically and culturally diverse languages and dialects to study how mLLMs resolve idiomatic ambiguity across high-, mid-, and low-resource settings; (ii) we provide idioms grounded in both literal and figurative sentence contexts as well as multi-turn dialogues; and (iii) we benchmark current LLMs and reveal persistent gaps in low-resource languages and for literal readings; while conversational context often improves accuracy, it does not close these gaps, highlighting persistent limitations in multilingual idiom understanding; (iv) we additionally show that activation steering along memorization and reasoning dimensions yields consistent gains, especially for low-resource languages, with MMLU-Pro \cite{wang2024mmlupro} directions transferring effectively to MIDI, suggesting a shared mechanism behind the memorization–reasoning tradeoff.

\begin{figure*}[t!]
    \centering
\includegraphics[width=0.9\textwidth]{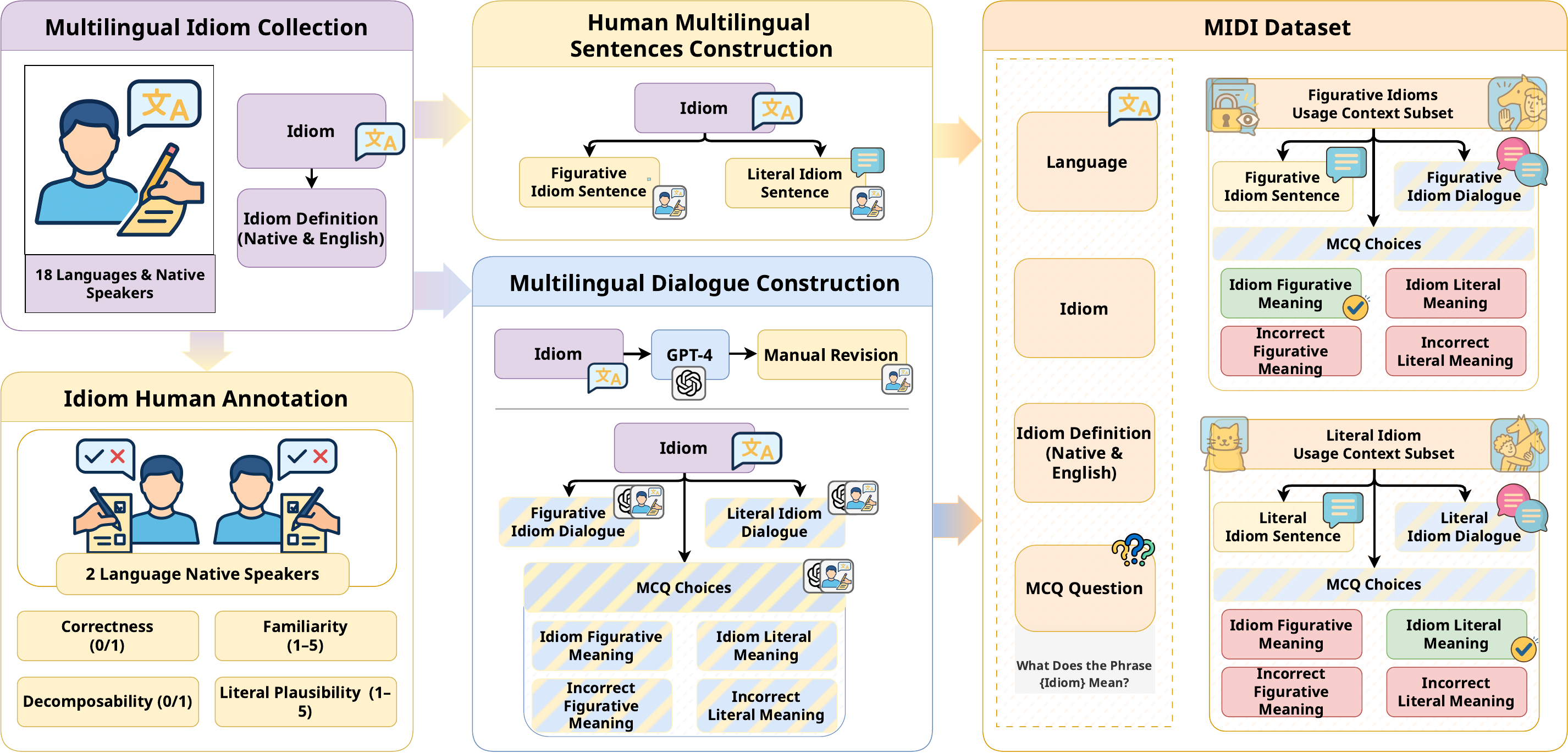}
    \caption{\textbf{MIDI construction pipeline.} Native speakers collect and annotate idioms (18 languages) with bilingual definitions, then create figurative/literal sentence contexts and LLM-generated (manually revised) dialogues with MCQ options, producing paired figurative and literal usage subsets.}
    \label{fig:dataset_placeholder}
    \setlength{\abovecaptionskip}{3pt}
    \setlength{\belowcaptionskip}{0pt}
\end{figure*}

\section{Related Work}

A significant body of literature has introduced various idiom datasets, reflecting a growing interest in the computational modeling of figurative language. \texttt{MAGPIE} \cite{haagsma-etal-2020-magpie} is one of the most popular datasets for English providing 56K instances of potentially idiomatic expressions across 1.7K idioms, focusing primarily on sentence-level annotations without multilingual extension. The more recent \texttt{DICE} dataset \cite{mi-etal-2025-rolling} introduces a contrastive evaluation framework for idiomatic expressions also for English only, examining whether models can correctly interpret idioms in context. 

\citet{lai-etal-2023-multilingual} propose multilingual multi-figurative language detection using a unified prompt-based framework, demonstrating generalization across languages, figures of speech, and zero-shot settings. \citet{khoshtab-etal-2025-comparative} compare how different large language models interpret idioms and similes across multiple languages ---especially including new Persian datasets--- and how various prompting strategies affect their performance.
Beyond idioms, proverb interpretation has served as a lens for examining cultural knowledge and reasoning in multilingual language models. \texttt{MAPS} \cite{liu-etal-2024-multilingual} introduces a multicultural dataset of proverbs and sayings across six languages in conversational contexts, providing analyses of memorization versus reasoning and highlighting cultural gaps in mLLMs.

Prior work also explores the binary classification of literal versus figurative usage \cite{tedeschi-etal-2022-id10m,de-luca-fornaciari-etal-2024-hard}, establishing the importance of context for idiom interpretation with evaluations restricted to English or a small set of languages.

Table \ref{tab:datasets} summarizes the comparison of the existing datasets~\cite{moussallem-etal-2018-lidioms, haagsma-etal-2020-magpie, tayyar-madabushi-etal-2021-astitchinlanguagemodels-dataset, tedeschi-etal-2022-id10m, kim-etal-2025-memorization} with \textbf{MIDI} along 5 dimensions: (1) \textit{Number of languages:} \textbf{MIDI} covers the largest and most diverse set of languages to date, (2) \textit{MCQ availability}: only one of 5 works provides multiple-choice questions for evaluating idioms, (3) \textit{Context Granularity:} This dimension assesses the contextual granularity of the dataset, specifically whether it provides sentence-level or conversation-level information, (4) \textit{Low Resource:} this dimension relates to whether the languages that are considered low-resource, and (5) \textit{Multi-dimensional idiom Annotation:} this checks if additional human annotations (beyond simple definitions) related to the idiom are available; in \textbf{MIDI} we include 3 additional psycholinguistic properties of an idiom.

\section{Dataset Construction}
\label{sec:dataset}

\begin{table}[t]
    \centering
    \footnotesize
    \setlength{\tabcolsep}{5pt}
    \renewcommand{\arraystretch}{0.65}
    \setlength{\abovecaptionskip}{3pt}
    \setlength{\belowcaptionskip}{0pt}
    \begin{tabularx}{\linewidth}{@{}Xccc@{}}
    \toprule
    \textbf{Language} & \textbf{\#Idioms} & \textbf{\#Literal (\%)} & \textbf{\#Contexts} \\
    \midrule
    \multicolumn{4}{@{}l@{}}{\cellcolor{highres}\textbf{High-Resource}} \\
    Chinese        & 300 & 100 & 1200 \\
    Japanese       & 115 & 95  & 448 \\
    Russian        & 213 & 98  & 842 \\

    \multicolumn{4}{@{}l@{}}{\cellcolor{medres}\textbf{Mid-Resource}} \\
    Arabic (UAE)   & 100 & 98  & 396 \\
    Indonesian     & 108 & 57  & 330 \\
    Vietnamese     & 100 & 0   & 200 \\

    \multicolumn{4}{@{}l@{}}{\cellcolor{lowres}\textbf{Low-Resource}} \\
    Arabic (Egypt)   & 100 & 84  & 368 \\
    Arabic (Morocco) & 99  & 94  & 384 \\
    Arabic (Syria)   & 100 & 100 & 400 \\
    Persian          & 102 & 80  & 368 \\
    Javanese         & 104 & 92  & 400 \\
    Kannada          & 198 & 100 & 792 \\
    Kazakh           & 100 & 100 & 400 \\
    Minangkabau      & 100 & 49  & 298 \\
    Sundanese        & 100 & 72  & 344 \\
    Tamil            & 99  & 96  & 388 \\
    Telugu           & 139 & 99  & 552 \\
    Yoruba           & 101 & 33  & 268 \\
    \midrule
    \textbf{Total / Avg.} & \textbf{2,278} & \textbf{80} & \textbf{8,378} \\
    \bottomrule
    \end{tabularx}
    \caption{Languages in \textbf{MIDI} by resource tier (\colorbox{highres}{High}/\colorbox{medres}{Mid}/\colorbox{lowres}{Low}). \#Literal is the share of figurative idioms with a literal counterpart; \#Contexts counts sentence and dialogue instances (figurative+literal).}
    \label{tab:languages}
\end{table}

The process depicted in Figure \ref{fig:dataset_placeholder} starts with multilingual idiom collection across 18 languages and dialects, carried out by native speakers with backgrounds in natural language processing and related fields, from different age groups, both male and female. The languages included in the dataset and their statistics are shown in Table~\ref{tab:languages}. For each language, annotators collected idioms that are commonly used and familiar to native speakers.

\textbf{MIDI} contains 2,278 idioms and 8,378 usage contexts, with each idiom appearing in both as a sentence and a dialogue. On average, 80\% of the idioms have a literal counterpart, though this varies across languages. Some languages show a high overlap between figurative and literal meanings, while others do not. In particular, Vietnamese idioms in the dataset are exclusively non-literal, and low-resource languages such as Yoruba and Minangkabau contain relatively few idioms with literal interpretations. 

\paragraph{Dimension Annotation.}
Each idiom is validated by at least two native speakers and annotated along several dimensions following \citet{LibbenTitone2008}: decomposability (binary), familiarity (1--5), and literal plausibility (1--5).

\paragraph{Figurative and Literal Example Sentences.} The collected idioms are supported by two example sentences: one illustrating the idiom figurative usage and another demonstrating its literal, word-by-word interpretation, context. The sentences are manually authored (and also manually curated) and do not originate from existing sources, ensuring no data leakage into the training process.

\paragraph{Multilingual Dialogue Construction.} For each idiom, we asked GPT-5 to generate two types of dialogues: a dialogue with the idiom in figurative meaning and another with its literal meaning. The prompts for each dialogue type are presented in Appendix~\ref{appx:prompts}. In addition to the dialogues, we asked GPT-5 to generate multiple-choice questions (MCQs) to further probe idiom understanding. The answer options are the correct figurative meaning, the correct literal meaning, and corresponding incorrect distractors. All generated dialogues and MCQs undergo a manual revision step by the same annotators from previous step to ensure linguistic quality and semantic correctness. However, for Arabic (UAE), Minangkabau and Yoruba, the LLM-generated text was almost unusable, requiring annotators to rewrite nearly 100\% of the content from scratch.

\paragraph{Final dataset.} The collected data is organized into the multilingual idioms dataset \textbf{MIDI}. Each instance is explicitly associated with its language, idiom, and bilingual (native language and English) definition. Overall, \textbf{MIDI} consists of two complementary subsets: a figurative idiom usage context subset and a literal idiom usage context subset. Each subset contains the corresponding idiom sentence, idiom dialogue, and an MCQ asking \textit{``What does the phrase [IDIOM] mean?''}, along with answer choices. In the figurative subset, the figurative meaning is marked as correct, while in the literal subset, the literal meaning is marked as correct. Instances of the \textbf{MIDI} dataset alongside per-language statistics are provided in Appendix~\ref{appx:dataset_extra}.


\definecolor{highres}{HTML}{C8E6C9}    
\definecolor{medres}{HTML}{FFF9C4}     
\definecolor{lowres}{HTML}{FFCDD2}     
\definecolor{rowgray}{HTML}{F5F5F5}    

\begin{table*}[t]
\centering
\setlength{\tabcolsep}{5pt}
\setlength{\abovecaptionskip}{3pt}
\setlength{\belowcaptionskip}{0pt}
\renewcommand{\arraystretch}{0.8}
\small
\begin{tabular}{@{}l cc cc cc cc cc cc 
c@{}}
\toprule
& \multicolumn{6}{c}{\textbf{Idiom Usage Context}} & \multicolumn{6}{c}{\textbf{Idiom Usage Type}} & \\
\cmidrule(lr){2-7} \cmidrule(lr){8-13}
& \multicolumn{2}{c}{\cellcolor{highres}\textbf{High}} & \multicolumn{2}{c}{\cellcolor{medres}\textbf{Mid}} & \multicolumn{2}{c}{\cellcolor{lowres}\textbf{Low}} & \multicolumn{2}{c}{\cellcolor{highres}\textbf{High}} & \multicolumn{2}{c}{\cellcolor{medres}\textbf{Mid}} & \multicolumn{2}{c}{\cellcolor{lowres}\textbf{Low}} & \\
\textbf{Model} & \textbf{Sent} & \textbf{Conv} & \textbf{Sent} & \textbf{Conv} & \textbf{Sent} & \textbf{Conv} & \textbf{Fig} & \textbf{Lit} & \textbf{Fig} & \textbf{Lit} & \textbf{Fig} & \textbf{Lit} & \textbf{Overall} \\
\midrule
\textcolor{gray}{\textit{Random}} & \textcolor{gray}{\textit{25.0}} & \textcolor{gray}{\textit{25.0}} & \textcolor{gray}{\textit{25.0}} & \textcolor{gray}{\textit{25.0}} & \textcolor{gray}{\textit{25.0}} & \textcolor{gray}{\textit{25.0}} & \textcolor{gray}{\textit{25.0}} & \textcolor{gray}{\textit{25.0}} & \textcolor{gray}{\textit{25.0}} & \textcolor{gray}{\textit{25.0}} & \textcolor{gray}{\textit{25.0}} & \textcolor{gray}{\textit{25.0}} & \textcolor{gray}{\textit{25.0}} \\
\hdashline
\noalign{\vskip 1pt}
\multicolumn{14}{@{}l}{\textit{\small Proprietary}} \\
\rowcolor{rowgray}
GPT-5.2 & 65.1 & 67.1 & \textbf{90.0} & \textbf{91.1} & \underline{68.5} & \underline{77.1} & \underline{97.0} & 35.2 & \underline{97.2} & \underline{76.8} & \underline{86.7} & \underline{58.9} & \underline{74.4} \\
Gemini 2.5 Pro & \underline{67.5} & 71.2 & \underline{89.5} & \underline{90.9} & \textbf{69.5} & \textbf{78.2} & \textbf{98.6} & 40.2 & \textbf{99.2} & 72.1 & \textbf{94.8} & 52.9 & \textbf{75.5} \\ 
\hdashline
\noalign{\vskip 1pt}
\multicolumn{14}{@{}l}{\textit{\small Open-Source}} \\
DeepSeek-R1 (70B) & 48.8 & 54.2 & 53.6 & 57.5 & 34.6 & 40.5 & 55.1 & 47.9 & 57.5 & 46.4 & 42.6 & 32.6 & 44.0 \\
\rowcolor{rowgray}
Gemma-3 (27B) & \underline{67.5} & \underline{73.2} & 86.9 & 90.0 & 58.8 & 70.9 & 96.9 & 43.8 & 96.2 & 71.7 & 76.4 & 53.3 & 71.0 \\
Llama-3.1 (8B) & 62.4 & 68.6 & 27.6 & 31.7 & 39.6 & 45.0 & 82.0 & \underline{49.0} & 29.3 & 33.5 & 43.1 & 41.6 & 49.5 \\
\rowcolor{rowgray}
Llama-3.3 (70B) & 64.6 & 65.2 & 65.5 & 75.2 & 53.5 & 62.4 & 87.5 & 42.3 & 77.9 & 59.4 & 70.7 & 45.2 & 63.6 \\
Mixtral (8$\times$7B) & \textbf{69.4} & \textbf{76.9} & 59.6 & 75.6 & 42.6 & 53.2 & 79.4 & \textbf{67.0} & 57.9 & \textbf{80.7} & 35.7 & \textbf{60.1} & 57.5 \\
\rowcolor{rowgray}
Qwen-3 (4B) & 30.6 & 38.3 & 31.0 & 35.1 & 27.5 & 28.5 & 43.3 & 25.7 & 36.3 & 32.8 & 27.1 & 28.9 & 30.8 \\
Qwen-3 (30B-MoE) & 40.8 & 37.4 & 24.1 & 23.5 & 25.6 & 25.8 & 43.9 & 34.3 & 24.4 & 23.3 & 24.1 & 27.2 & 31.2 \\
\midrule
\rowcolor{rowgray}
\multicolumn{1}{@{}l}{\textit{Average}} & \textit{57.4} & \textit{61.4} & \textit{58.6} & \textit{63.4} & \textit{46.7} & \textit{53.5} & \textit{76.0} & \textit{42.8} & \textit{64.0} & \textit{55.2} & \textit{55.7} & \textit{44.5} & \textit{55.3} \\
\bottomrule
\noalign{\vskip 1pt}
\multicolumn{14}{c}{\small\textit{Sent = Sentence \quad\quad Conv = Conversation \quad\quad Fig = Figurative \quad\quad Lit = Literal}} \\
\end{tabular}
\caption{Model performance (accuracy \%) on idiom comprehension MCQ across usage contexts and types, stratified by language resource availability (\colorbox{highres}{High}, \colorbox{medres}{Mid}, \colorbox{lowres}{Low}). \textbf{Bold} indicates best; \underline{underline} indicates second-best.}
\label{tab:main_results}
\end{table*}

\section{Experiment}
Using \textbf{MIDI}, we benchmark multilingual LLMs to characterize idiom comprehension across (i) language resource levels, (ii) sentence vs.\ conversational contexts, and (iii) figurative vs.\ literal usage.

\subsection{Experimental Setup}
 We evaluate both proprietary and open-source LLMs representing the current state of the art in multilingual NLP. The proprietary models include GPT-5.2 \citep{openai2025gpt5card} and Gemini 2.5 Pro \citep{comanici2025gemini25pushingfrontier}, which represent frontier capabilities in multilingual reasoning and understanding. The open-source models include DeepSeek-R1-Distill-Llama (70B) \citep{deepseekai2025deepseekr1incentivizingreasoningcapability}, Gemma-3 (27B Instruct) \citep{gemmateam2025gemma3technicalreport}, Llama-3.1 (8B Instruct), Llama-3.3 (70B Instruct) \citep{grattafiori2024llama3herdmodels}, Mixtral (8$\times$7B Instruct) \citep{jiang2024mixtralexperts}, Qwen-3 (4B Instruct; 30B-MoE Instruct) \cite{yang2025qwen3technicalreport}. This set spans a range of sizes and architectures, enabling analysis of how scale and design choices relate to idiom understanding.

For evaluation we cast idiom comprehension (\textit{reasoning + memorization}) as a four-way MCQ task. Given an idiom embedded in context (sentence-level or multi-turn dialogue), the model selects the correct interpretation from four options: the figurative meaning, the literal meaning, and two distractors (one figurative, one literal). We report accuracy across idiom usage conditions (figurative or literal), with random baseline performance at 25\%. For exact task formulation, refer to the Appendix~\ref{app:prompt_default}.

All evaluations are conducted in a zero-shot setting to assess models' inherent capabilities without task-specific fine-tuning using \texttt{lm-evaluation-harness} \citep{eval-harness}.  We aggregate results across instances and analyze performance along three axes: (i) \textit{context type} (sentence vs.\ dialogue), (ii) \textit{usage type} (figurative vs.\ literal), and (iii) \textit{language resource availability level} (high-, medium-, and low-resource). This decomposition isolates the effects of context, semantic ambiguity, and data availability on idiom understanding.

\subsection{Idiom Comprehension MCQ}
\label{sec:main_experiment_mcq}
Table~\ref{tab:main_results} presents idiom comprehension results across all evaluation dimensions.

\paragraph{Overall Observation.} Proprietary models substantially outperform open-source alternatives: Gemini 2.5 Pro achieves the highest accuracy at 75\%, followed closely by GPT-5.2 at 74\%, while the best open-source model, Gemma-3 (27B), reaches 71\%. At the lower end, Qwen-3 (4B) and Qwen-3 (30B-MoE) perform near random chance both at 31\%, indicating that smaller or sparsely-activated architectures struggle with multilingual idiom understanding. Notably, DeepSeek-R1 (70B) underperforms relative to its parameter count, achieving only 44\% overall, substantially below the smaller Gemma-3, suggesting that model architecture and training objectives may matter as much as scale. An unexpected pattern emerges in the resource availability level results: medium-resource languages often yield the highest accuracy, GPT-5.2 achieves 90\% on mid-resource sentences versus 65\% on high-resource. We propose two possible explanations. The first concerns the quality of the pretraining data: mid-resource corpora may be more carefully filtered and curated than their larger, but noisier, high-resource counterparts. The second relates to the composition of our dataset and the nature of these languages: mid-resource languages have the lowest proportion of literal counterparts (e.g., Vietnamese at 0\% and Indonesian at 52.7\%), meaning their evaluation skews toward figurative usage mostly.

\paragraph{Resource Availability.} Performance degrades as language resource availability decreases. Averaging across models, high-resource languages achieve 59\% accuracy, medium-resource languages reach 61\%, and low-resource languages attain only 50\%. This degradation is particularly pronounced for open-source models; for instance, Gemma-3 (27B) achieves 70\% on high-resource languages but drops to 65\% on low-resource languages. The gap is somewhat narrower for proprietary models, with Gemini 2.5 Pro maintaining relatively robust performance across resource levels (69.4\% high, 90.2\% mid, 73.9\% low), suggesting that larger-scale training may partially mitigate resource scarcity effects. We also observe that medium-resource languages yield particularly strong performance for several models, GPT-5.2 and Gemma\mbox{-3} reach 90\% and 87\% respectively, on medium-resource sentences vs.\ 65\% and 68\% on high-resource sentences, indicating that resource tier is not the only determinant of difficulty.

\definecolor{highres}{HTML}{C8E6C9}    
\definecolor{medres}{HTML}{FFF9C4}     
\definecolor{lowres}{HTML}{FFCDD2}     
\definecolor{rowgray}{HTML}{F5F5F5}    

\begin{table}[t]
\centering
\setlength{\tabcolsep}{6pt}
\setlength{\belowcaptionskip}{0pt}
\renewcommand{\arraystretch}{0.5}
\footnotesize
\begin{tabular}{@{}l ccc c@{}}
\toprule
& \multicolumn{3}{c}{\textbf{Memorization}} & \\
\cmidrule(lr){2-4}
\textbf{Model} & \cellcolor{highres}\textbf{High} & \cellcolor{medres}\textbf{Mid} & \cellcolor{lowres}\textbf{Low} & \textbf{Overall} \\
\midrule
\textcolor{gray}{\textit{Random}} & \textcolor{gray}{\textit{25.0}} & \textcolor{gray}{\textit{25.0}} & \textcolor{gray}{\textit{25.0}} & \textcolor{gray}{\textit{25.0}} \\
\hdashline
\noalign{\vskip 1pt}
\multicolumn{5}{@{}l}{\textit{\small Proprietary}} \\
\rowcolor{rowgray}
GPT-5.2 & \underline{97.8} & \underline{94.4} & \underline{84.0} & \underline{89.3} \\
Gemini 2.5 Pro & \textbf{98.8} & \textbf{97.7} & \textbf{92.5} & \textbf{95.0} \\
\hdashline
\noalign{\vskip 1pt}
\multicolumn{5}{@{}l}{\textit{\small Open-Source}} \\
DeepSeek-R1 (70B) & 43.5 & 47.0 & 34.1 & 38.4 \\
\rowcolor{rowgray}
Gemma-3 (27B) & 96.4 & 86.9 & 67.3 & 78.9 \\
Llama-3.1 (8B) & 71.4 & 29.2 & 43.4 & 51.4 \\
\rowcolor{rowgray}
Llama-3.3 (70B) & 38.6 & 44.9 & 56.9 & 51.4 \\
Mixtral (8$\times$7B) & 76.8 & 53.6 & 43.4 & 55.4 \\
\rowcolor{rowgray}
Qwen-3 (4B) & 50.3 & 30.2 & 23.7 & 33.3 \\
Qwen-3 (30B-MoE) & 49.7 & 24.2 & 22.4 & 33.4 \\
\midrule
\rowcolor{rowgray}
\textit{Average} & \textit{69.3} & \textit{56.4} & \textit{52.0} & \textit{58.5} \\
\bottomrule
\noalign{\vskip 1pt}
\multicolumn{5}{c}{\small\textit{\colorbox{highres}{High}, \colorbox{medres}{Mid}, \colorbox{lowres}{Low} = Resource Level}} \\
\end{tabular}
\caption{Model performance (accuracy \%) on idiom memorization (figurative meaning identification given no context). \textbf{Bold} = best; \underline{underline} = second-best.}
\label{tab:memorization}
\end{table}

\paragraph{Context Type.} Conversational contexts consistently improve model performance compared to sentence-level contexts. Across all models and resource availability levels, conversation-based evaluation yields an average accuracy of 59\% compared to 54\% for sentences. This improvement is most pronounced for low-resource languages, where the gap between conversation (54\%) and sentence (47\%) contexts reaches 7 percentage points. This is consistent with the idea that multi-turn discourse provides additional pragmatic cues including speaker intentions and discourse coherence, that help models disambiguate idiomatic usage even when lexical familiarity is limited.

\begin{figure*}[t]
    \centering
    \scalebox{1}[1]{%
        \includegraphics[width=1\linewidth]{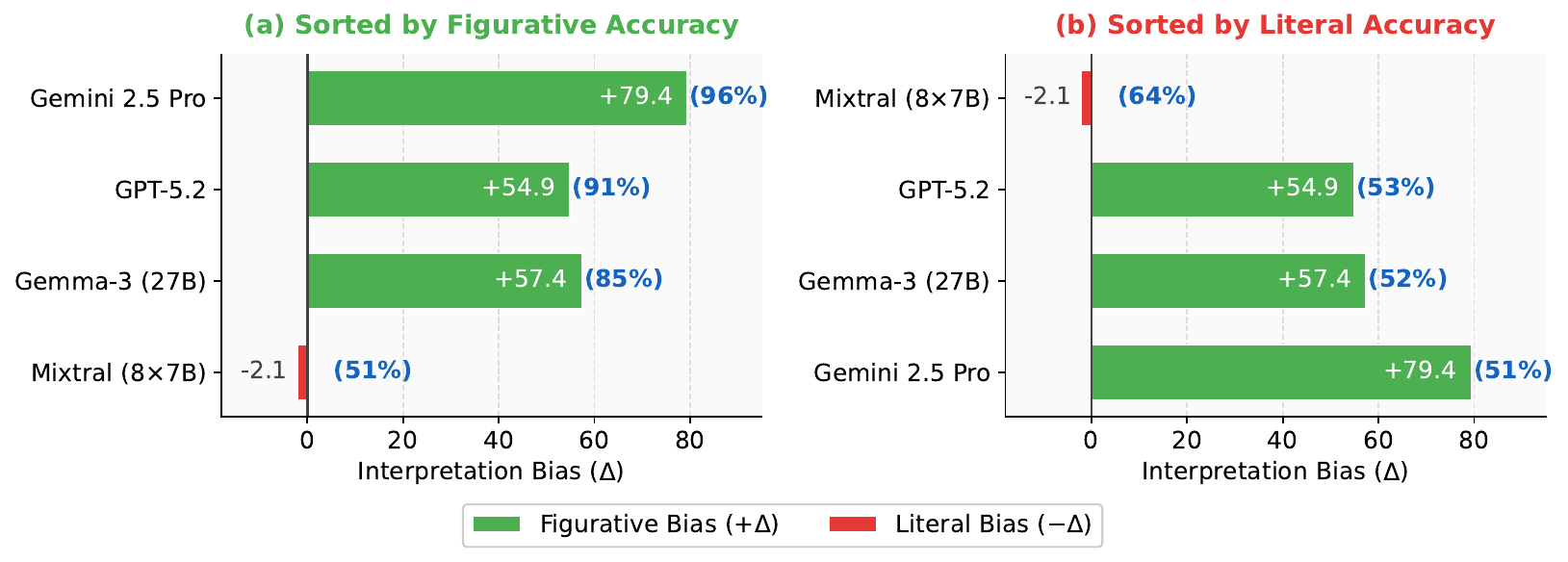}%
    }
    \caption{Interpretation bias ($\Delta$ = figurative--literal preference, \%) vs.\ idiom comprehension accuracy: models sorted by (a) figurative and (b) literal accuracy (parentheses show accuracies \%).}
    \label{fig:bias_correlation}
\end{figure*}

\paragraph{Figurative vs. Literal Comprehension.} Models exhibit substantially higher accuracy on figurative usage interpretation (65\% average) compared to literal usage (48\% average). This asymmetry is most extreme for high-resource languages, where the gap reaches 33 percentage points (76\% figurative vs. 43\% literal). The pattern suggests that models more readily identify figurative meanings, likely due to stronger training signal from canonical idiom definitions, but struggle to recognize when contextual cues indicate a compositional literal reading. This highlights a fundamental limitation in current models' ability to flexibly disambiguate idiom usage based on context. Mixtral (8$\times$7B) is a notable exception, achieving the highest literal accuracy across all resource availability levels (67\% high, 81\% mid, 60\% low), indicating stronger sensitivity to literal-signaling context, though at the cost of lower figurative performance in medium and low-resource settings.

These findings show that idiom comprehension remains challenging for current LLMs, particularly in low-resource languages and literal usages. A detailed breakdown by language and evaluation condition is provided in Appendix~\ref{appx:main_results_extend}. We next analyze the underlying mechanisms: Section~\ref{sec:memorization_and_reasoning} isolates memorization and contextual reasoning, Section~\ref{sec:bias} examines models’ interpretation bias toward figurative readings, Section~\ref{sec:steering} studies how steering memorization and reasoning affects performance, and Section~\ref{sec:human} highlights the gap between human and model performance in idiom comprehension across a subset of languages.

\definecolor{highres}{HTML}{C8E6C9}    
\definecolor{medres}{HTML}{FFF9C4}     
\definecolor{lowres}{HTML}{FFCDD2}     
\definecolor{rowgray}{HTML}{F5F5F5}    

\begin{table}[t]
\centering
\setlength{\tabcolsep}{6pt}
\setlength{\belowcaptionskip}{0pt}
\renewcommand{\arraystretch}{0.5} 
\footnotesize
\begin{tabular}{@{}l ccc c@{}}
\toprule
& \multicolumn{3}{c}{\textbf{Reasoning}} & \\
\cmidrule(lr){2-4}
\textbf{Model} & \cellcolor{highres}\textbf{High} & \cellcolor{medres}\textbf{Mid} & \cellcolor{lowres}\textbf{Low} & \textbf{Overall} \\
\midrule
\textcolor{gray}{\textit{Random}} & \textcolor{gray}{\textit{25.0}} & \textcolor{gray}{\textit{25.0}} & \textcolor{gray}{\textit{25.0}} & \textcolor{gray}{\textit{25.0}} \\
\hdashline
\noalign{\vskip 1pt}
\multicolumn{5}{@{}l}{\textit{\small Proprietary}} \\
\rowcolor{rowgray}
GPT-5.2 & \underline{98.2} & 98.5 & \underline{95.6} & \underline{96.4} \\
Gemini 2.5 Pro & \textbf{98.9} & \textbf{99.7} & \textbf{97.5} & \textbf{98.0} \\
\hdashline
\noalign{\vskip 1pt}
\multicolumn{5}{@{}l}{\textit{\small Open-Source}} \\
DeepSeek-R1 (70B) & 54.3 & 61.3 & 53.9 & 54.7 \\
\rowcolor{rowgray}
Gemma-3 (27B) & 97.5 & \underline{98.9} & 91.2 & 93.9 \\
Llama-3.1 (8B) & 96.8 & 54.3 & 73.7 & 78.1 \\
\rowcolor{rowgray}
Llama-3.3 (70B) & 95.9 & 91.0 & 77.8 & 85.2 \\
Mixtral (8$\times$7B) & 94.5 & 91.0 & 72.2 & 80.8 \\
\rowcolor{rowgray}
Qwen-3 (4B) & 44.3 & 36.8 & 31.7 & 36.9 \\
Qwen-3 (30B-MoE) & 65.5 & 35.9 & 31.3 & 43.0 \\
\midrule
\rowcolor{rowgray}
\textit{Average} & \textit{82.9} & \textit{74.1} & \textit{69.4} & \textit{74.1} \\
\bottomrule
\noalign{\vskip 1pt}
\multicolumn{5}{c}{\small\textit{\colorbox{highres}{High}, \colorbox{medres}{Mid}, \colorbox{lowres}{Low} = Resource Level}} \\
\end{tabular}
\caption{Model performance (accuracy \%) on idiom reasoning (figurative meaning identification given usage context and English meaning). \textbf{Bold} = best; \underline{underline} = second-best.}
\label{tab:reasoning}
\end{table}

\section{Analysis}

\subsection{Memorization and Reasoning}
\label{sec:memorization_and_reasoning} 

\noindent\textbf{Memorization.} The memorization settings evaluates whether a model can recall the figurative meaning of an idiom without any contextual support. In this task, the model is presented with an idiom in isolation and asked \textit{"What is the figurative meaning of this idiom?"}, forcing it to rely solely on its stored knowledge rather than contextual reasoning (see Appendix~\ref{app:prompts} for details). Results are shown in Table~\ref{tab:memorization}. Proprietary models achieve consistently high performance across all resource levels. This suggests broad memorization of idiomatic expressions across languages. In contrast, open-source models exhibit substantial variation. Larger models such as Gemma-3 and Mixtral show relatively strong memorization in high-resource languages, while smaller models and those evaluated on mid- and low-resource languages perform substantially worse. This pattern mirrors the main results, where limited memorization often coincides with weaker figurative understanding in context.

\noindent\textbf{Reasoning.} In contrast to memorization, the reasoning setting examines the model's ability to identify the figurative meaning of idiom when both the usage context and the English meaning are provided. In this task, the model model is no longer required to rely on memorization knowledge alone, but instead must use the given information to reason about the correct interpretation (see Appendix~\ref{app:prompts} for prompt details). Results are shown in Table~\ref{tab:reasoning}. Proprietary models achieve near-perfect accuracy across all resource levels, indicating strong and consistent reasoning ability even in low-resource languages. Meanwhile, open-source models benefit substantially from the availability of context and meaning, with several large models, such as Gemma-3 and Mixtral, approaching proprietary performance in high-resource settings, though performance still degrades in mid- and low-resource languages. Compared to memorization, reasoning reduces but does not eliminate the performance gap between proprietary and open-source models, suggesting that context helps compensate for limited idiomatic knowledge but does not fully resolve it. Performance patterns are largely consistent between sentence and conversational contexts, with detailed results reported in Table~\ref{tab:appendix_reasoning_context}.

\subsection{Interpretation Bias}
\label{sec:bias}
To understand the asymmetric figurative-literal performance gap observed in Section~\ref{sec:main_experiment_mcq}, we analyze models' default interpretation tendencies when presented with ambiguous idiom prompts lacking contextual cues. We query each model with the prompt \textit{``What does the phrase \{idiom\} mean?''} followed by the figurative and literal meanings as options, without any usage context to guide interpretation (see Appendix~\ref{app:prompts} for details).

Figure~\ref{fig:bias_correlation} reveals a strong correlation between interpretation bias and task performance. Models exhibiting strong figurative bias, preferring figurative interpretations in ambiguous contexts, achieve substantially higher accuracy on figurative meaning identification. Gemini 2.5 Pro shows the strongest figurative bias ($\Delta$=+79.4) and achieves the highest figurative accuracy (96\%), followed by GPT-5.2 ($\Delta$=+54.9, 91\%) and Gemma-3 ($\Delta$=+57.4, 85\%). This pattern suggests that models' strong figurative performance stems partly from a default tendency to interpret idioms figuratively, likely reinforced by training data where idioms predominantly appear with figurative meanings.

Conversely, the literal accuracy ranking reveals an inverse pattern: models with weaker figurative bias achieve better literal comprehension. Mixtral, with near-neutral bias ($\Delta$=--2.1), achieves the highest literal accuracy (64\%). Among the strong figurative-bias models, GPT-5.2 (53\%) and Gemma-3 (52\%) outperform Gemini 2.5 Pro (51\%) on literal tasks---notably, their literal performance ordering mirrors their bias magnitude, with Gemini's strongest figurative bias ($\Delta$=+79.4) corresponding to the weakest literal comprehension despite its overall superiority on figurative interpretation. This suggests that excessive figurative bias impairs models' ability to recognize literal usage, even for otherwise highly capable models.

Full bias analysis across all models is provided in Appendix~\ref{appx:bias_extend}.


\definecolor{tierhigh}{HTML}{C8E6C9} 
\definecolor{tiermid}{HTML}{FFF9C4}  
\definecolor{tierlow}{HTML}{FFCDD2}

\begin{table}[t]
\centering
\small
\setlength{\tabcolsep}{5pt}
\renewcommand{\arraystretch}{1}
\resizebox{\linewidth}{!}{%
\begin{tabular}{llc|cccc}
\toprule
Model & Vector Source & Type &
\cellcolor{tierhigh}\textbf{High} &
\cellcolor{tiermid}\textbf{Mid} &
\cellcolor{tierlow}\textbf{Low} &
Avg.\\
\midrule
\multirow{4}{*}{Llama-3.1 (8B)} &
\multirow{2}{*}{MIDI-derived} & Mem.  & 0.25 & 0.37 & 0.64 & 0.44 \\
& & Reas. & 0.14 & 0.34 & 0.60 & 0.36 \\
& \multirow{2}{*}{MMLU-Pro}     & Mem.  & 0.22 & 0.32 & 0.59 & 0.40 \\
& & Reas. & 0.20 & 0.31 & 0.63 & 0.38 \\
\midrule
\multirow{4}{*}{Qwen-3 (4B)} &
\multirow{2}{*}{MIDI-derived} & Mem.  & 0.52 & 0.34 & 0.86 & 0.64 \\
& & Reas. & 0.32 & 0.28 & 0.65 & 0.40 \\
& \multirow{2}{*}{MMLU-Pro}     & Mem.  & 0.71 & 0.41 & 1.26 & 0.78 \\
& & Reas. & 0.38 & 0.36 & 0.62 & 0.41 \\
\bottomrule
\end{tabular}%
}
\caption{Best-layer steering gains ($\Delta$ accuracy in percentage points), macro-averaged across languages within each resource tier and subsequently averaged over all tasks.}
\label{tab:steering_delta_summary}
\end{table}

\subsection{Activation Steering Analysis}
\label{sec:steering}

\paragraph{Setup.}
To better understand the interplay between memorization and reasoning in LLMs, we extend our analysis to activation steering, examining how latent representations can generalize and enhance performance in low-resource settings. We perform inference-time \emph{activation steering} by linearly modifying residual-stream activations at a selected layer using activation addition (ActAdd)~\citep{turner2023activationaddition}.
Following the Linear Reasoning Features (LiReFs) framework~\citep{hong2025reasoningmemorization}, we derive a steering direction via a difference-in-means estimator.
Given prompt sets eliciting reasoning ($D_{\textsc{reas}}$) and memorization ($D_{\textsc{mem}}$), let $h^{(\ell)}(x)$ denote the residual-stream activation at layer~$\ell$ for the final prompt position.
We compute:
\begin{equation}
    \small
    r^{(\ell)} = \frac{1}{|D_{\textsc{reas}}|}\sum_{x \in D_{\textsc{reas}}} h^{(\ell)}(x) \;-\; \frac{1}{|D_{\textsc{mem}}|}\sum_{x \in D_{\textsc{mem}}} h^{(\ell)}(x).
    \label{eq:diff_means}
\end{equation}
At inference time, we steer by adding a scaled copy of this direction vector:
\begin{equation}
    h^{(\ell)}(x) \leftarrow h^{(\ell)}(x) + \alpha r^{(\ell)}, \quad \text{with } |\alpha| = 0.1.
    \label{eq:actadd}
\end{equation}

We compare steering directions from two sources: \textbf{MIDI-derived vectors}, computed from MIDI prompts, and \textbf{MMLU-Pro vectors}, extracted from an independent benchmark~\citep{wang2024mmlupro}.
For each source, we evaluate two steering polarities: \emph{memorization steering} ($\alpha<0$) and \emph{reasoning steering} ($\alpha>0$).

\paragraph{Tasks and layer sweep.}
We evaluate steering across the five MIDI configurations introduced earlier: \textbf{Dialogue Standard}, \textbf{Dialogue Reasoning}, \textbf{Memorization}, \textbf{Sentence Standard}, and \textbf{Sentence Reasoning}.
For computational tractability, we sweep alternate layers beginning at layer~3 ($\ell \in \{3,5,7,\dots\}$; through $\ell{=}31$ for Llama-3.1~8B and $\ell{=}35$ for Qwen-3~4B) and report results at the \textbf{best-performing layer} selected independently for each combination of language, task, steering type, and vector source.

\paragraph{Main findings.}
Steering yields \textbf{modest but consistent} accuracy gains, with the largest improvements concentrated in \textbf{low-resource languages} (Appendix~\ref{appx:steering_details}), consistent with a headroom effect.
MMLU-Pro vectors transfer well to idiom understanding, performing comparably to MIDI-derived vectors overall and slightly better for Qwen-3~4B in aggregate.

Interestingly, memorization steering often yields slightly larger gains than reasoning steering (Table~\ref{tab:steering_delta_summary}).
This supports the broader picture from our controlled probes: memorization and reasoning are \textbf{not cleanly separable}.
For idioms, ``memorization'' features can act as \emph{sense anchors} (a stored inventory of candidate meanings) that \emph{enable} downstream contextual reasoning and disambiguation rather than competing with it.

Steering is also \textbf{layer-sensitive}: the best layer varies across languages and tasks, and suboptimal layers can degrade performance.
Notably, the distribution of best layers is similar between MIDI-derived and MMLU-Pro vectors within each model; see Figure~\ref{fig:best_layer_hist} in Appendix~\ref{appx:steering_details}, suggesting that both sources tap into a shared internal component mediating the memorization--reasoning tradeoff, rather than dataset-specific artifacts.

Table~\ref{tab:steering_delta_summary} reports best-layer gains, macro-averaged across languages within each resource tier (to control for tier size) and then averaged over all five tasks.
Full per-task $\Delta$-accuracy grids, direction norms, and flip-rate diagnostics are provided in Appendix~\ref{appx:steering_details}.

\definecolor{tierlow}{HTML}{E3242B} 

\begin{table}[t]
\centering
\small
\setlength{\tabcolsep}{6pt}
\renewcommand{\arraystretch}{1.1}
\resizebox{\linewidth}{!}{%
\begin{tabular}{@{}l c c c@{}}
\toprule
\textbf{Language}
& \textbf{Human}
& \textbf{Best Proprietary}
& \textbf{Best Open-Source} \\
\midrule
\multicolumn{4}{@{}l}{\cellcolor{highres}\textbf{High-Resource Languages}} \\
\rowcolor{rowgray}
Japanese         &  88 &  71 \textcolor{tierlow}{($-$17)} &  69 \textcolor{tierlow}{($-$19)} \\
Russian          &  99 &  65 \textcolor{tierlow}{($-$34)} &  86 \textcolor{tierlow}{($-$13)} \\
\midrule
\multicolumn{4}{@{}l}{\cellcolor{medres}\textbf{Mid-Resource Languages}} \\
\rowcolor{rowgray}
Arabic (UAE)     &  97 &  87 \textcolor{tierlow}{($-$10)} &  81 \textcolor{tierlow}{($-$16)} \\
Indonesian       &  91 &  88 \textcolor{tierlow}{($-$3)}  &  88 \textcolor{tierlow}{($-$3)}  \\
\rowcolor{rowgray}
Vietnamese       & 100 & 100 (0)                        & 100 (0) \\
\midrule
\multicolumn{4}{@{}l}{\cellcolor{lowres}\textbf{Low-Resource Languages}} \\
\rowcolor{rowgray}
Arabic (Egypt)   &  92 &  83 \textcolor{tierlow}{($-$9)}  &  75 \textcolor{tierlow}{($-$17)} \\
Arabic (Morocco) &  89 &  62 \textcolor{tierlow}{($-$27)} &  65 \textcolor{tierlow}{($-$24)} \\
\rowcolor{rowgray}
Minangkabau      &  96 &  82 \textcolor{tierlow}{($-$14)} &  63 \textcolor{tierlow}{($-$33)} \\
Yoruba           &  96 &  89 \textcolor{tierlow}{($-$7)}  &  23 \textcolor{tierlow}{($-$73)} \\
\midrule \rowcolor{rowgray}
\textbf{Average}
& \textbf{94}
& \textbf{81} \textcolor{tierlow}{\textbf{($-$13)}}
& \textbf{72} \textcolor{tierlow}{\textbf{($-$22)}} \\
\bottomrule
\end{tabular}%
}
\caption{Comparison of human accuracy (\%) on a 10\% random sample of idioms against the best-performing proprietary and open-source models for each language. The value in parentheses shows the model's gap from human accuracy in percentage points (\textcolor{tierlow}{red} for below-human performance). Additional details in Table~\ref{tab:human_vs_models}.}
\label{tab:analysis_human_vs_models}
\end{table}

\subsection{Human--Model Gap}
\label{sec:human}
To verify that \textbf{MIDI}'s idiom comprehension task setup (Section~\ref{sec:main_experiment_mcq}; Appendix~\ref{app:prompt_default}) is clear and interpretable for native speakers, and to establish a reference point against which model performance can be contextualized, we conduct a human evaluation on a 10\% random subset of idioms from nine languages (half of \textbf{MIDI}) spanning all three resource tiers. For each language, we also report the best-performing proprietary and open-source model performance on the same subset, enabling a matched comparison that uses, for each language, the best model in each category.

Table~\ref{tab:analysis_human_vs_models} presents the results per-language. Human annotators achieve an overall accuracy of 94\%, with individual language scores ranging from 88\% to 100\%. Both model categories fall well short of this reference: the best proprietary model in each language averages 81\% ($-$13 points), while the best open-source model per language averages 72\% ($-$22 points). Humans also outperform the stronger of the two model categories in eight out of nine languages, with parity only in Vietnamese (100\%). The gaps are substantial across languages, with especially large shortfalls for open-source models in low-resource settings. The most extreme case is Yoruba ($-$73), where the best open-source model drops to nearly random accuracy, whereas the best proprietary model retains 89\% on the same samples.

These findings indicate that the \textbf{MIDI} idiom comprehension task is interpretable and can be reliably completed by native speakers under the same setup used for models, implying that model failure cannot be attributed to ambiguity in the task itself. They further underscore that there is still considerable room for improvement for current LLMs on idiom understanding, especially for low-resource languages and for open-source models. Detailed per-condition results and a consistency check against overall dataset performance are provided in Appendix~\ref{app:human_eval}.

\section{Conclusion}
We introduce \textbf{MIDI}, a multilingual idiom understanding benchmark covering 18 languages and dialects across high-, medium-, and low-resource tiers, with idioms presented in sentence and dialogue contexts and, where possible, in both figurative and literal forms. Zero-shot evaluations reveal that even state-of-the-art models struggle with idiom comprehension, with performance dropping sharply in low-resource languages and literal interpretations proving harder than figurative ones. Conversational context improves accuracy, particularly in low-resource settings, but does not fully close these gaps, highlighting persistent limitations in multilingual idiom understanding.

Our controlled probes show that reasoning with explicit definitions improves accuracy but does not fully close the gap. {Memorization} strongly correlates with successful figurative interpretation. Together, these findings suggest a hybrid view: {memorization} supplies candidate senses and priors, while {reasoning} leverages context to select the correct meaning. This explains why memorization aids reasoning and why these capacities are intertwined. Consistent with this entanglement, models exhibit a strong {figurative bias}, where priors boosting figurative accuracy can hinder literal comprehension when compositional interpretation is needed.

We further show that {activation steering} along memorization and reasoning dimensions yields modest yet consistent gains, especially in low-resource languages. Directions from {MMLU-Pro} transfer effectively to MIDI, and optimal layer distributions overlap, indicating a shared mechanism behind the memorization–reasoning tradeoff rather than dataset-specific artifacts. Finally, a human evaluation on a subset covering half of \textbf{MIDI}'s languages shows that native speakers can consistently solve the idiom comprehension task, whereas both proprietary and open-source models fall substantially short, with the performance gap for open-source models increasing significantly in low-resource languages.





\section*{Limitations}

\paragraph{Scope of Memorization Vs.\ Reasoning Analysis.}
Our controlled experiments, which aim to disentangle memorization from contextual reasoning, focus only on \textit{figurative} interpretations. The reasoning task asks, \textit{"What is the figurative meaning of \texttt{[IDIOM]}?"}, while the memorization task directly provides the figurative definition. Thus, our framework does not explicitly assess how memorization and reasoning interact for \textit{literal} readings, where the model must suppress or override the usually preferred idiomatic meaning.

\paragraph{Uneven Coverage and Comparability Across Languages.}
Languages differ in whether idioms admit plausible literal counterparts (e.g., Vietnamese idioms in our dataset are purely figurative). Moreover, \textbf{MIDI} is imbalanced across resource tiers (12 low-resource vs.\ 3 medium and 3 high) and in the number of idioms/contexts per language. Thus, despite its size and diversity, \textbf{MIDI} is better suited for identifying broad tier-level and literal–figurative patterns than for tightly controlled, language-matched comparisons.

\paragraph{Agreement for Psycholinguistic Ratings.}
Although each instance was reviewed by at least two native speakers to verify linguistic quality and answer correctness, but we do not yet report inter-annotator agreement for familiarity, decomposability, or literal plausibility. These additional ratings should therefore be treated as informative but potentially noisy, rather than as definitive gold-standard measures.


\paragraph{Dialogue Naturalness.}
Dialogue contexts are first generated by an LLM and then manually revised by native speakers to ensure semantic and linguistic accuracy. This yields coherent, well‑controlled conversations but may underrepresent the stylistic and pragmatic variability found in naturally occurring interactions, slightly limiting naturalistic coverage.

\paragraph{Cross-Context Option Consistency.}
The MCQ options were created during the dialogue revision phase, rather than being developed together with the previously written sentence contexts. For idioms that support multiple plausible literal readings (especially in Japanese), the selected literal option was chosen to fit the dialogue, but may not perfectly align with the literal interpretation suggested by the sentence itself. This mismatch may help explain why sentence-context accuracy is lower than dialogue-based accuracy. Ensuring strict consistency of options across both context types is left for future dataset revisions.

\paragraph{MCQ Evaluation Constraints.}
We evaluate idiom comprehension using multiple-choice questions with manually curated distractors, allowing controlled comparisons across languages and settings. However, MCQs cannot fully capture open-ended interpretation or generation, so extending \textbf{MIDI} to free-form explanations and downstream tasks is left for future work.

\section*{Ethical Considerations}
All human-authored data were manually reviewed to ensure the absence of harmful, offensive, or inappropriate content. Annotators provided informed consent for their contributions to be used and distributed for research purposes and were compensated through co-authorship or through fair compensation. No sensitive or personally identifiable information was collected or disclosed, the dataset does not involve vulnerable populations, and the study poses minimal risk to participants.

\section*{Acknowledgments}

We thank Ibrahim Alsarraj for his help in checking the quality of some Arabic (Syria) data, Dhahi A. for annotating Arabic (UAE) instances, Fajar Mohamad Ridwan for reviewing Sundanese data, and Anass Baatite for reviewing Moroccan instances. We thank Mai Shaaban for her help with dimension annotation for Egyptian Arabic data. We acknowledge the assistance of Sahaja Nayak, Prathibha M. Shetty and Pushpalatha B. K. in annotating and reviewing the Kannada instances, Shion Hara for annotating the Japanese instances, Olanrewaju Taofiq for annotating the Yoruba instances, and Denis Bolshakov, Alexey Matyushin,  Anna Azarova, and Alexandra Shestakova for annotating the Russian instances in the dataset.


\bibliography{custom}

\clearpage
\appendix
\twocolumn

\section{Data Statements for MIDI}
\label{appx:data-statement}

\subsection{General Information}

\noindent\textbf{Dataset title:} \textbf{MIDI}

\noindent\textbf{Dataset version:} 1.0 (November 2025)

\noindent\textbf{Data statement version:} 1.0 (December 2025)





\subsection{Executive Summary}

MIDI is a multilingual dataset for evaluating idiom understanding across languages with different resource levels. It covers 18 languages and dialects across high-, mid-, and low-resource tiers and contains 2,278 idioms instantiated in 8,378 usage contexts. Each idiom appears in both sentence-level and dialogue-level contexts and is paired with a multiple-choice question designed to test idiom comprehension under realistic discourse conditions. When a literal counterpart exists, both figurative and literal usages are included, with an average literal coverage of 80\% across languages. The dataset supports controlled evaluation of idiom comprehension as well as diagnostic analyses that isolate memorization and reasoning using consistent prompt templates (Appendix~\ref{app:prompts}).

\subsection{Curation Rationale}

MIDI was curated to enable systematic analysis of idiom understanding in multilingual settings, particularly for languages that are underrepresented in existing benchmarks. Prior datasets often focus on a small number of high-resource languages or evaluate idioms in isolation, which limits their ability to capture how idioms are interpreted in context. MIDI addresses this gap by including a broad set of languages and by embedding idioms in both sentences and short dialogues. The inclusion of both figurative and literal usages further allows evaluation of how models distinguish idiomatic meaning from compositional interpretations.

\subsection{Documentation for Source}

Idioms were collected by native-speaker annotators for each language based on their linguistic knowledge and familiarity with common usage. The dataset does not rely on a single predefined lexicon or corpus. Example sentences were manually authored or curated to illustrate natural idiom usage. Dialogue contexts were initially generated using an LLM and then carefully reviewed and edited by native speakers to ensure linguistic quality and semantic correctness. All content was created specifically for this dataset to reduce the risk of data leakage and to maintain consistency across languages.

\subsection{Language Variety}

The dataset includes 18 languages and dialects spanning multiple language families, scripts, and typological properties. Languages are grouped into high-resource (Chinese, Japanese, Russian), mid-resource (Indonesian, Vietnamese, and the UAE dialect of Arabic), and low-resource categories (Javanese, Persian, Kannada, Telugu, Tamil, Minangkabau, Sundanese, Kazakh, Yoruba, and Arabic dialects from Syria, Egypt, and Morocco). Both standard and widely used regional varieties are included where appropriate, allowing analysis across different levels of linguistic resource availability.



\section{Annotation Guidelines}
\label{appx:annotation-guidelines}

As described in Section~\ref{sec:dataset}, dialogues and multiple-choice questions were initially generated by an LLM and subsequently revised by native-speaker annotators. The revision guidelines are two-fold, covering both the dialogues and the multiple-choice options.

\paragraph{Dialogue Revision.} Annotators were instructed to review and, where necessary, revise or rewrite each LLM-generated conversation to ensure it reads as a natural, fluent, and culturally appropriate spoken exchange. This involved paying close attention to matching the tone, vocabulary, and style of each speaker, and confirming that the idiom is used correctly in context---whether figuratively or literally. Where the generated output was insufficiently natural or culturally appropriate, annotators were expected to rewrite the content entirely.

\paragraph{Option Revision.} Each MCQ consists of four answer choices: the correct figurative meaning, the correct literal meaning, and two distractors (one figurative, one literal). A strict JSON structure was required for all answer options. Annotators ensured that the correct option accurately reflects the intended meaning of the idiom as used in the dialogue (figurative for figurative-usage dialogues, literal for literal-usage dialogues). The incorrect options were required to be plausible and similar in length to the correct answer, with distractors drawing on the inverse literal/figurative interpretation as well as other misleading but wrong readings, so as to prevent elimination by surface cues alone.

\section{Dialogue Generation Prompts}
\label{appx:prompts}

\tcbset{
  promptbox/.style={
    enhanced,
    colback=gray!3,
    colframe=gray!55,
    boxrule=2pt,
    arc=2mm,
    left=2mm,right=2mm,top=1.2mm,bottom=1.2mm,
    fonttitle=\bfseries,
    coltitle=gray!30!black,
    breakable
  }
}

\begin{figure}[ht!]
    \centering
    \begin{tcolorbox}[promptbox,breakable=false,title={GPT prompt for figurative dialogue generation}]
    \ttfamily\footnotesize
Idiom: [IDIOM]\\
Idiom meaning: [IDIOM MEANING IN ENGLISH]\\[2pt]
Generate a short 3-turn dialogue in which the final utterance includes the idiom
``[IDIOM]'' in [LANGUAGE]. After the dialogue, include the question:
``Does [IDIOM] imply a figurative or literal meaning? What does it mean?''\\
Provide a multiple-choice question with three options. The correct answer is:
[IDIOM MEANING IN NON-ENGLISH]. Create two incorrect options.\\
Output format (JSON):\\
\{\\
\ \ "conversation": "The full 3-turn dialogue",\\
\ \ "question": "Does [IDIOM] imply a figurative or literal meaning? What does it mean?",\\
\ \ "correct option": "[IDIOM MEANING IN NON-ENGLISH]",\\
\ \ "incorrect option 1": "The idiom literal meaning in non English",\\
\ \ "incorrect option 2": "The other incorrect meaning with regard to its figurative meaning",\\
\ \ "incorrect option 3": "The other incorrect meaning with regard to its literal meaning"\\
\}
    \end{tcolorbox}
    \caption{Prompt used to generate dialogues where the idiom is intended to be interpreted figuratively.}
    \label{fig:fig_prompt}
\end{figure}

\begin{figure}[ht!]
    \centering
    \begin{tcolorbox}[promptbox,title={GPT prompt for literal dialogue generation}]
    \ttfamily\footnotesize
Phrase: [IDIOM]\\
Generate a short 3-turn dialogue in which the final utterance includes the phrase
``[IDIOM]'' in [LANGUAGE], used with its literal meaning. After the dialogue,
include the question: ``Does [IDIOM] imply a figurative or literal meaning? What does it mean?''\\
Provide a multiple-choice question with three options. The correct answer is: [IDIOM LITERAL MEANING IN NON-ENGLISH]. Create two incorrect options.\\
Output format (JSON):\\
\{\\
\ \ "conversation": "The full 3-turn dialogue",\\
\ \ "question": "Does [IDIOM] imply a figurative or literal meaning? What does it mean?",\\
\ \ "correct option": "[IDIOM LITERAL MEANING IN NON-ENGLISH]",\\
\ \ "incorrect option 1": "[IDIOM FIGURATIVE MEANING IN NON-ENGLISH]",\\
\ \ "incorrect option 2": "The other incorrect meaning with regard to its figurative meaning",\\
\ \ "incorrect option 3": "The other incorrect meaning with regard to its literal meaning"\\
\}
    \end{tcolorbox}
    \caption{Prompt used to generate dialogues where the idiom is intended to be interpreted literally.}
    \label{fig:lit_prompt}
\end{figure}

\section{Prompt Templates}
\label{app:prompts}

This appendix lists the exact prompt templates used across all idiom comprehension settings. 
All prompts instantiate the placeholders \texttt{\{context\}}, \texttt{\{idiom\}}, \texttt{\{options\}}, and (when applicable) \texttt{\{idiom\_meaning\}}.

\subsection{Default Idiom Comprehension task (Reasoning + Memorization)}
\label{app:prompt_default}
This is our main evaluation setting used in Section~\ref{sec:main_experiment_mcq}. The model receives the usage context (sentence or dialogue) and selects one out of four options: the figurative meaning, the literal meaning, and two distractors (one figurative, one literal). The correct choice depends on how the idiom usage type: if it is used literally, the answer reflects its direct, dictionary meaning; if it is used figuratively, the answer reflects its non-literal, symbolic meaning.

\begin{tcolorbox}[promptbox,title={Reasoning + Memorization Prompt}]
\ttfamily\footnotesize
You are tasked with selecting the most appropriate option based on the context provided below.

Context: \{context\}

What does the phrase "\{idiom\}" mean?

Options:
{options}
\end{tcolorbox}

\subsection{Reasoning Isolation (Context + Meaning Hint)}
\label{app:prompt_reasoning}
To reduce reliance on definitional recall, we provide the idiom’s meaning as a hint and evaluate whether models still select the correct answer conditioned on discourse context. This is used in reasoning task in Section~\ref{sec:memorization_and_reasoning}.

\begin{tcolorbox}[promptbox,title={Reasoning Isolation Prompt}]
\ttfamily\footnotesize
You are tasked with selecting the most appropriate option based on the context provided below.

Context: \{context\}

"\{idiom\}" means \{idiom\_meaning\}

What does the phrase "\{idiom\}" mean?

Options:
\{options\}
\end{tcolorbox}

\subsection{Memorization Isolation (No Context)}
\label{app:prompt_memorization}
This setting removes all usage context, probing definitional knowledge of idioms. The model must select the figurative meaning from the same four options. This is used in memorization task in Section~\ref{sec:memorization_and_reasoning}.

\begin{tcolorbox}[promptbox,title={Memorization Isolation Prompt}]
\ttfamily\footnotesize
You are tasked with selecting the most appropriate option based on the context provided below.

What is the figurative meaning of "\{idiom\}"?

Options:
\{options\}
\end{tcolorbox}

\subsection{Idiom Interpretation Bias}
\label{app:prompt_bias}
We measure interpretation bias using a binary choice format (A\/B). In this setup, \texttt{\{options\}} contains exactly two correct answers (e.g., figurative vs.\ literal), formatted as choices A and B. This is used in bias analysis in Section~\ref{sec:bias}.

\begin{tcolorbox}[promptbox,title={Bias Prompt (Binary A/B)}]
\ttfamily\footnotesize
You are tasked with selecting the most appropriate option based on the context provided below.

What does the phrase "\{idiom\}" mean?

Options:
\{options\}
\end{tcolorbox}

\paragraph{Option formatting.}
For the 4-way MCQ settings, \texttt{\{options\}} is formatted as four labeled choices (A--D). For the bias setting, \texttt{\{options\}} is formatted as two labeled choices (A--B). We keep the question surface form fixed across settings whenever possible to avoid confounding effects from prompt formulation.

\section{Detailed Main Results}
\label{appx:main_results_extend}

In addition to the main results presented in the paper, Tables~\ref{tab:appendix_context} and ~\ref{tab:appendix_type} provide detailed breakdowns of model performance on the idiom comprehension task (reasoning + memorization). A detailed summary of the overall performance of the models in the different languages included in our evaluation is provided in Table~\ref{tab:per_language_accuracy}.

\definecolor{highres}{HTML}{C8E6C9}    
\definecolor{medres}{HTML}{FFF9C4}     
\definecolor{lowres}{HTML}{FFCDD2}     
\definecolor{rowgray}{HTML}{F5F5F5}    

\begin{table}[!t]
\centering
\setlength{\tabcolsep}{5pt}
\renewcommand{\arraystretch}{1.15}
\small
\resizebox{\columnwidth}{!}{%
\begin{tabular}{@{}l ccc ccc c@{}}
\toprule
& \multicolumn{3}{c}{\textbf{Sentence}} & \multicolumn{3}{c}{\textbf{Conversation}} & \\
\cmidrule(lr){2-4} \cmidrule(lr){5-7}
\textbf{Model} & \cellcolor{highres}\textbf{High} & \cellcolor{medres}\textbf{Mid} & \cellcolor{lowres}\textbf{Low} & \cellcolor{highres}\textbf{High} & \cellcolor{medres}\textbf{Mid} & \cellcolor{lowres}\textbf{Low} & \textbf{Overall} \\
\midrule
\textcolor{gray}{\textit{Random}} & \textcolor{gray}{\textit{25.0}} & \textcolor{gray}{\textit{25.0}} & \textcolor{gray}{\textit{25.0}} & \textcolor{gray}{\textit{25.0}} & \textcolor{gray}{\textit{25.0}} & \textcolor{gray}{\textit{25.0}} & \textcolor{gray}{\textit{25.0}} \\
\hdashline
\noalign{\vskip 2pt}
\multicolumn{8}{@{}l}{\textit{\small Proprietary}} \\
\rowcolor{rowgray}
GPT-5.2 & \underline{65.1} & \textbf{90.0} & \underline{68.5} & 67.1 & \textbf{91.1} & \underline{77.1} & \underline{74.4} \\
Gemini 2.5 Pro & \textbf{67.5} & \underline{89.5} & \textbf{69.5} & 71.2 & \underline{90.9} & \textbf{78.2} & \textbf{75.5} \\
\hdashline
\noalign{\vskip 2pt}
\multicolumn{8}{@{}l}{\textit{\small Open-Source}} \\
DeepSeek-R1 (70B) & 48.8 & 53.6 & 34.6 & 54.2 & 57.5 & 40.5 & 44.0 \\
\rowcolor{rowgray}
Gemma-3 (27B) & \textbf{67.5} & 86.9 & 58.8 & \textbf{73.2} & 90.0 & 70.9 & 71.0 \\
Llama-3.1 (8B) & 62.4 & 27.6 & 39.6 & 68.6 & 31.7 & 45.0 & 49.5 \\
\rowcolor{rowgray}
Llama-3.3 (70B) & 64.6 & 65.5 & 53.5 & 65.2 & 75.2 & 62.4 & 63.6 \\
Mixtral (8$\times$7B) & 69.4 & 59.6 & 42.6 & 76.9 & 75.6 & 53.2 & 57.5 \\
\rowcolor{rowgray}
Qwen-3 (4B) & 30.6 & 31.0 & 27.5 & 38.3 & 35.1 & 28.5 & 30.8 \\
Qwen-3 (30B-MoE) & 40.8 & 24.1 & 25.6 & 37.4 & 23.5 & 25.8 & 31.2 \\
\midrule
\rowcolor{rowgray}
\multicolumn{1}{@{}l}{\textit{Average}} & \textit{57.4} & \textit{58.6} & \textit{46.7} & \textit{61.4} & \textit{63.4} & \textit{53.5} & \textit{55.3} \\
\bottomrule
\noalign{\vskip 3pt}
\multicolumn{8}{c}{\small\textit{\colorbox{highres}{High} = High Resource \quad\quad \colorbox{medres}{Mid} = Mid Resource \quad\quad \colorbox{lowres}{Low} = Low Resource}} \\
\end{tabular}
}
\caption{Model performance (accuracy \%) on idiom comprehension by usage context (Sentence vs.\ Conversation), stratified by language resource availability. \textbf{Bold} indicates best; \underline{underline} indicates second-best.}
\label{tab:appendix_context}
\end{table}
\definecolor{highres}{HTML}{C8E6C9}    
\definecolor{medres}{HTML}{FFF9C4}     
\definecolor{lowres}{HTML}{FFCDD2}     
\definecolor{rowgray}{HTML}{F5F5F5}    

\begin{table}[!t]
\centering
\setlength{\tabcolsep}{5pt}
\renewcommand{\arraystretch}{1.15}
\small
\resizebox{\columnwidth}{!}{%
\begin{tabular}{@{}l ccc ccc c@{}}
\toprule
& \multicolumn{3}{c}{\textbf{Figurative}} & \multicolumn{3}{c}{\textbf{Literal}} & \\
\cmidrule(lr){2-4} \cmidrule(lr){5-7}
\textbf{Model} & \cellcolor{highres}\textbf{High} & \cellcolor{medres}\textbf{Mid} & \cellcolor{lowres}\textbf{Low} & \cellcolor{highres}\textbf{High} & \cellcolor{medres}\textbf{Mid} & \cellcolor{lowres}\textbf{Low} & \textbf{Overall} \\
\midrule
\textcolor{gray}{\textit{Random}} & \textcolor{gray}{\textit{25.0}} & \textcolor{gray}{\textit{25.0}} & \textcolor{gray}{\textit{25.0}} & \textcolor{gray}{\textit{25.0}} & \textcolor{gray}{\textit{25.0}} & \textcolor{gray}{\textit{25.0}} & \textcolor{gray}{\textit{25.0}} \\
\hdashline
\noalign{\vskip 2pt}
\multicolumn{8}{@{}l}{\textit{\small Proprietary}} \\
\rowcolor{rowgray}
GPT-5.2 & \underline{97.0} & \underline{97.2} & \underline{86.7} & 35.2 & \underline{76.8} & \underline{58.9} & \underline{74.4} \\
Gemini 2.5 Pro & \textbf{98.6} & \textbf{99.2} & \textbf{94.8} & 40.2 & 72.1 & 52.9 & \textbf{75.5} \\
\hdashline
\noalign{\vskip 2pt}
\multicolumn{8}{@{}l}{\textit{\small Open-Source}} \\
DeepSeek-R1 (70B) & 55.1 & 57.5 & 42.6 & 47.9 & 46.4 & 32.6 & 44.0 \\
\rowcolor{rowgray}
Gemma-3 (27B) & 96.9 & 96.2 & 76.4 & 43.8 & 71.7 & 53.3 & 71.0 \\
Llama-3.1 (8B) & 82.0 & 29.3 & 43.1 & \underline{49.0} & 33.5 & 41.6 & 49.5 \\
\rowcolor{rowgray}
Llama-3.3 (70B) & 87.5 & 77.9 & 70.7 & 42.3 & 59.4 & 45.2 & 63.6 \\
Mixtral (8$\times$7B) & 79.4 & 57.9 & 35.7 & \textbf{67.0} & \textbf{80.7} & \textbf{60.1} & 57.5 \\
\rowcolor{rowgray}
Qwen-3 (4B) & 43.3 & 36.3 & 27.1 & 25.7 & 32.8 & 28.9 & 30.8 \\
Qwen-3 (30B-MoE) & 43.9 & 24.4 & 24.1 & 34.3 & 23.3 & 27.2 & 31.2 \\
\midrule
\rowcolor{rowgray}
\multicolumn{1}{@{}l}{\textit{Average}} & \textit{76.0} & \textit{64.0} & \textit{55.7} & \textit{42.8} & \textit{55.2} & \textit{44.5} & \textit{55.3} \\
\bottomrule
\noalign{\vskip 3pt}
\multicolumn{8}{c}{\small\textit{\colorbox{highres}{High} = High Resource \quad\quad \colorbox{medres}{Mid} = Mid Resource \quad\quad \colorbox{lowres}{Low} = Low Resource}} \\
\end{tabular}%
}
\caption{Model performance (accuracy \%) on idiom comprehension by usage type (Figurative vs.\ Literal), stratified by language resource availability. \textbf{Bold} indicates best; \underline{underline} indicates second-best.}
\label{tab:appendix_type}
\end{table}
\definecolor{highres}{HTML}{C8E6C9}    
\definecolor{medres}{HTML}{FFF9C4}     
\definecolor{lowres}{HTML}{FFCDD2}     
\definecolor{rowgray}{HTML}{F5F5F5}    

\begin{table*}[!t]
\centering
\setlength{\tabcolsep}{3pt}
\renewcommand{\arraystretch}{1.15}
\small
\resizebox{\textwidth}{!}{%
\begin{tabular}{@{}l ccc ccc cccccccccccc c@{}}
\toprule

 & \multicolumn{3}{c}{\cellcolor{highres}\textbf{High}}
 & \multicolumn{3}{c}{\cellcolor{medres}\textbf{Mid}}
 & \multicolumn{12}{c}{\cellcolor{lowres}\textbf{Low}}
 &  \\
\cmidrule(lr){2-4} \cmidrule(lr){5-7} \cmidrule(lr){8-19}
 \textbf{Model} & Zh & Ja & Ru
 & Ar-UAE & Id & Vi
 & Ar-EG & Ar-MA & Ar-SY & Fa & Jv & Kn & Kk & Min & Su & Ta & Te & Yo
 & \textbf{Overall}\\
\midrule
\textcolor{gray}{\textit{Random}}
 & \textcolor{gray}{\textit{25.0}}
 & \textcolor{gray}{\textit{25.0}}
 & \textcolor{gray}{\textit{25.0}}
 & \textcolor{gray}{\textit{25.0}}
 & \textcolor{gray}{\textit{25.0}}
 & \textcolor{gray}{\textit{25.0}}
 & \textcolor{gray}{\textit{25.0}}
 & \textcolor{gray}{\textit{25.0}}
 & \textcolor{gray}{\textit{25.0}}
 & \textcolor{gray}{\textit{25.0}}
 & \textcolor{gray}{\textit{25.0}}
 & \textcolor{gray}{\textit{25.0}}
 & \textcolor{gray}{\textit{25.0}}
 & \textcolor{gray}{\textit{25.0}}
 & \textcolor{gray}{\textit{25.0}}
 & \textcolor{gray}{\textit{25.0}}
 & \textcolor{gray}{\textit{25.0}}
 & \textcolor{gray}{\textit{25.0}}
 & \textcolor{gray}{\textit{25.0}} \\
\hdashline
\noalign{\vskip 2pt}
\multicolumn{20}{@{}l}{\textit{\small Proprietary}} \\
\rowcolor{rowgray}
GPT-5.2 & 71.2 & 65.4 & 63.1 & \textbf{87.1} & \underline{90.3} & \underline{97.0} & \textbf{84.8} & \textbf{70.2} & \underline{70.3} & 66.0 & \textbf{77.3} & \textbf{88.7} & \textbf{63.8} & \underline{73.2} & \underline{73.5} & \textbf{79.6} & \textbf{77.4} & \underline{52.2} & \underline{74.4} \\
Gemini 2.5 Pro & 71.5 & \textbf{70.8} & 67.1 & \underline{83.6} & \textbf{91.2} & \textbf{98.5} & \underline{82.1} & 64.8 & 69.8 & 64.1 & \underline{75.3} & \underline{87.8} & \underline{63.5} & \textbf{85.9} & \textbf{78.2} & 77.3 & 73.0 & \textbf{79.5} & \textbf{75.5} \\
\hdashline
\noalign{\vskip 2pt}
\multicolumn{20}{@{}l}{\textit{\small Open-Source}} \\
\rowcolor{rowgray}
DeepSeek-R1 (70B) & 50.3 & 48.4 & 55.9 & 42.4 & 58.2 & 68.0 & 48.9 & 35.5 & 43.0 & 56.3 & 33.5 & 44.0 & 41.3 & 30.2 & 27.9 & 36.5 & 29.3 & 25.4 & 44.0 \\
Gemma-3 (27B) & \textbf{72.9} & \underline{69.9} & 69.4 & 81.6 & 89.4 & \textbf{98.5} & 80.2 & \underline{66.6} & \textbf{70.5} & \underline{67.9} & 72.5 & 81.3 & 60.5 & 48.7 & 51.7 & 78.1 & \underline{75.0} & 23.9 & 71.0 \\
\rowcolor{rowgray}
Llama-3.1 (8B) & 68.3 & 67.6 & 61.2 & 33.3 & 32.4 & 23.0 & 42.1 & 44.8 & 37.8 & 51.1 & 31.8 & 65.6 & 33.0 & 29.5 & 28.2 & 63.3 & 51.4 & 25.4 & 49.5 \\
Llama-3.3 (70B) & 54.3 & \underline{69.9} & \underline{71.5} & 82.3 & 64.8 & 69.5 & 78.5 & \underline{66.6} & 66.0 & \textbf{68.2} & 45.8 & 81.8 & \underline{63.5} & 23.8 & 28.2 & \underline{78.8} & 72.6 & 23.5 & 63.6 \\
\rowcolor{rowgray}
Mixtral (8$\times$7B) & \underline{71.8} & 67.6 & \textbf{80.2} & 56.3 & 72.4 & 70.5 & 58.2 & 55.4 & 62.0 & 59.5 & 41.3 & 57.0 & 38.5 & 32.9 & 34.3 & 59.7 & 36.2 & 23.1 & 57.5 \\
Qwen-3 (4B) & 35.0 & 34.2 & 34.7 & 31.3 & 47.9 & 23.5 & 29.9 & 27.2 & 31.5 & 30.2 & 29.8 & 27.3 & 25.3 & 24.5 & 22.7 & 27.8 & 32.4 & 23.9 & 30.8 \\
\rowcolor{rowgray}
Qwen-3 (30B-MoE) & 64.3 & 25.7 & 27.3 & 23.2 & 26.4 & 23.0 & 23.4 & 24.6 & 23.5 & 23.4 & 27.5 & 26.3 & 24.3 & 25.5 & 27.6 & 24.5 & 30.4 & 23.9 & 31.2 \\
\midrule
\multicolumn{1}{@{}l}{\textit{Average}} & \textit{62.2} & \textit{57.7} & \textit{58.9} & \textit{57.9} & \textit{63.7} & \textit{63.5} & \textit{58.7} & \textit{50.6} & \textit{52.7} & \textit{54.1} & \textit{48.3} & \textit{62.2} & \textit{46.0} & \textit{41.6} & \textit{41.4} & \textit{58.4} & \textit{53.1} & \textit{33.4} & \textit{55.3} \\
\bottomrule
\noalign{\vskip 3pt}
\multicolumn{20}{c}{\small\textit{\colorbox{highres}{High} = High Resource \quad\quad \colorbox{medres}{Mid} = Mid Resource \quad\quad \colorbox{lowres}{Low} = Low Resource}} \\
\end{tabular}%
}
\caption{Model performance (accuracy \%) on idiom comprehension across 18 languages, stratified by resource availability. \textbf{Bold} indicates best; \underline{underline} indicates second-best. \textbf{Language codes:} \small{Zh = Chinese, Ja = Japanese, Ru = Russian, Ar-UAE/EG/MA/SY = Arabic (UAE / Egypt / Morocco / Syria), Id = Indonesian, Vi = Vietnamese, Fa = Persian, Jv = Javanese, Kn = Kannada, Kk = Kazakh, Min = Minangkabau, Su = Sundanese, Ta = Tamil, Te = Telugu, Yo = Yoruba.}}
\label{tab:per_language_accuracy}
\end{table*}

\section{Reasoning Evaluation Details}
\label{appx:reasoning_extend}
\definecolor{highres}{HTML}{C8E6C9}    
\definecolor{medres}{HTML}{FFF9C4}     
\definecolor{lowres}{HTML}{FFCDD2}     
\definecolor{rowgray}{HTML}{F5F5F5}    

\begin{table}[!t]
\centering
\setlength{\tabcolsep}{5pt}
\renewcommand{\arraystretch}{1.15}
\small
\resizebox{\columnwidth}{!}{%
\begin{tabular}{@{}l ccc ccc c@{}}
\toprule
& \multicolumn{3}{c}{\textbf{Sentence}} & \multicolumn{3}{c}{\textbf{Conversation}} & \\
\cmidrule(lr){2-4} \cmidrule(lr){5-7}
\textbf{Model} & \cellcolor{highres}\textbf{High} & \cellcolor{medres}\textbf{Mid} & \cellcolor{lowres}\textbf{Low} & \cellcolor{highres}\textbf{High} & \cellcolor{medres}\textbf{Mid} & \cellcolor{lowres}\textbf{Low} & \textbf{Overall} \\
\midrule
\textcolor{gray}{\textit{Random}} & \textcolor{gray}{\textit{25.0}} & \textcolor{gray}{\textit{25.0}} & \textcolor{gray}{\textit{25.0}} & \textcolor{gray}{\textit{25.0}} & \textcolor{gray}{\textit{25.0}} & \textcolor{gray}{\textit{25.0}} & \textcolor{gray}{\textit{25.0}} \\
\hdashline
\noalign{\vskip 2pt}
\multicolumn{8}{@{}l}{\textit{\small Proprietary}} \\
\rowcolor{rowgray}
GPT-5.2 & \underline{98.4} & \underline{98.7} & \underline{94.7} & \underline{98.0} & 98.3 & \underline{96.4} & \underline{96.4} \\
Gemini 2.5 Pro & \textbf{99.1} & \textbf{99.7} & \textbf{96.5} & \textbf{98.8} & \textbf{99.7} & \textbf{98.5} & \textbf{98.0} \\
\hdashline
\noalign{\vskip 2pt}
\multicolumn{8}{@{}l}{\textit{\small Open-Source}} \\
DeepSeek-R1 (70B) & 53.0 & 59.6 & 52.0 & 55.6 & 63.0 & 55.8 & 54.7 \\
\rowcolor{rowgray}
Gemma-3 (27B) & 97.8 & \underline{98.7} & 91.3 & 97.1 & \underline{99.0} & 91.1 & 93.9 \\
Llama-3.1 (8B) & 97.4 & 55.4 & 72.8 & 96.2 & 53.2 & 74.5 & 78.1 \\
\rowcolor{rowgray}
Llama-3.3 (70B) & 96.2 & 88.2 & 76.4 & 95.6 & 93.8 & 79.3 & 85.2 \\
Mixtral (8$\times$7B) & 94.0 & 89.9 & 72.6 & 95.1 & 92.2 & 71.8 & 80.8 \\
\rowcolor{rowgray}
Qwen-3 (4B) & 44.7 & 38.5 & 30.7 & 44.0 & 35.0 & 32.8 & 36.9 \\
Qwen-3 (30B-MoE) & 66.9 & 35.6 & 30.5 & 64.1 & 36.1 & 32.1 & 43.0 \\
\midrule
\rowcolor{rowgray}
\multicolumn{1}{@{}l}{\textit{Average}} & \textit{83.1} & \textit{73.8} & \textit{68.6} & \textit{82.7} & \textit{74.5} & \textit{70.3} & \textit{74.1} \\
\bottomrule
\noalign{\vskip 3pt}
\multicolumn{8}{c}{\small\textit{\colorbox{highres}{High} = High Resource \quad\quad \colorbox{medres}{Mid} = Mid Resource \quad\quad \colorbox{lowres}{Low} = Low Resource}} \\
\end{tabular}
}
\caption{Model performance (accuracy \%) on idiom reasoning by usage context (Sentence vs.\ Conversation), evaluating figurative meaning identification when provided with usage context and English meaning. Stratified by language resource availability. \textbf{Bold} indicates best; \underline{underline} indicates second-best.}
\label{tab:appendix_reasoning_context}
\end{table}
\paragraph{Sentence vs.\ conversation in reasoning-only.}
Table~\ref{tab:appendix_reasoning_context} breaks down the reasoning-only setting (Appendix~\ref{app:prompt_reasoning}) by context type. Differences between sentence and dialogue contexts are small across models and resource tiers (typically within 1--2 points on average), with proprietary models remaining near ceiling in both settings. This suggests that once the idiom’s English meaning is provided, performance is largely insensitive to whether the usage context is a single sentence or a short dialogue.

\section{Interpretation Bias Details}
\label{appx:bias_extend}

When presented with ambiguous idiom prompts lacking contextual cues (``What does the phrase \textit{\{idiom\}} mean?''), models exhibit systematic interpretation biases. Table~\ref{tab:interpretation_bias} reports interpretation tendencies alongside weighted overall accuracy on figurative ($n$=2,278) and literal ($n$=1,952) meaning identification tasks, i.e Section~\ref{sec:main_experiment_mcq}.

\definecolor{highres}{HTML}{C8E6C9}    
\definecolor{medres}{HTML}{FFF9C4}     
\definecolor{lowres}{HTML}{FFCDD2}     
\definecolor{rowgray}{HTML}{F5F5F5}    

\begin{table}[!ht]
\centering
\setlength{\tabcolsep}{5pt}
\renewcommand{\arraystretch}{1.15}
\footnotesize
\resizebox{\columnwidth}{!}{%
\begin{tabular}{@{}l cc c cc@{}}
\toprule
& \multicolumn{2}{c}{\textbf{Interpretation (\%)}} & & \multicolumn{2}{c}{\textbf{Overall Accuracy}} \\
\cmidrule(lr){2-3} \cmidrule(lr){5-6}
\textbf{Model} & \cellcolor{highres}\textbf{Fig.} & \cellcolor{lowres}\textbf{Lit.} & \textbf{$\Delta$} & \cellcolor{highres}\textbf{Fig.} & \cellcolor{lowres}\textbf{Lit.} \\
\midrule
\textcolor{gray}{\textit{Random}} & \textcolor{gray}{\textit{50.0}} & \textcolor{gray}{\textit{50.0}} & \textcolor{gray}{\textit{0.0}} & \textcolor{gray}{\textit{25.0}} & \textcolor{gray}{\textit{25.0}} \\
\hdashline
\noalign{\vskip 2pt}
\multicolumn{6}{@{}l}{\textit{\small Proprietary}} \\
\rowcolor{rowgray}
Gemini 2.5 Pro & \textbf{89.7} & 10.3 & \textbf{79.4} & \textbf{96.4} & 50.8 \\
GPT-5.2 & 77.4 & 22.6 & 54.9 & \underline{90.9} & \underline{53.2} \\
\hdashline
\noalign{\vskip 2pt}
\multicolumn{6}{@{}l}{\textit{\small Open-Source}} \\
\rowcolor{rowgray}
DeepSeek-R1 (70B) & 47.5 & 52.5 & -5.0 & 48.0 & 38.8 \\
Gemma-3 (27B) & \underline{78.7} & 21.3 & \underline{57.4} & 84.7 & 52.1 \\
\rowcolor{rowgray}
Llama-3.1 (8B) & 53.9 & 46.1 & 7.8 & 52.0 & 43.1 \\
Llama-3.3 (70B) & 66.1 & 33.9 & 32.3 & 76.3 & 45.7 \\
\rowcolor{rowgray}
Mixtral (8$\times$7B) & 48.9 & \textbf{51.1} & -2.1 & 50.8 & \textbf{64.4} \\
Qwen-3 (4B) & 66.7 & 33.3 & 33.4 & 32.8 & 28.3 \\
\rowcolor{rowgray}
Qwen-3 (30B-MoE) & 57.4 & 42.6 & 14.7 & 29.6 & 29.1 \\
\midrule
\textit{Average} & \textit{65.2} & \textit{34.8} & \textit{30.3} & \textit{62.4} & \textit{45.1} \\
\bottomrule
\noalign{\vskip 2pt}
\multicolumn{6}{c}{\small\textit{$\Delta$ = Figurative -- Literal bias; \colorbox{highres}{Fig.} = Figurative, \colorbox{lowres}{Lit.} = Literal}} \\
\end{tabular}
}
\caption{Model interpretation bias and reasoning accuracy on ambiguous idiom prompts. \textit{Left}: Interpretation tendency when no context is provided (positive $\Delta$ = figurative preference). \textit{Right}: Weighted overall accuracy on figurative ($n$=2,278) and literal ($n$=1,952) meaning identification. \textbf{Bold} = highest; \underline{underline} = second-highest.}
\label{tab:interpretation_bias}
\end{table}

\begin{figure}[ht]
    \centering
    \includegraphics[width=1\columnwidth]{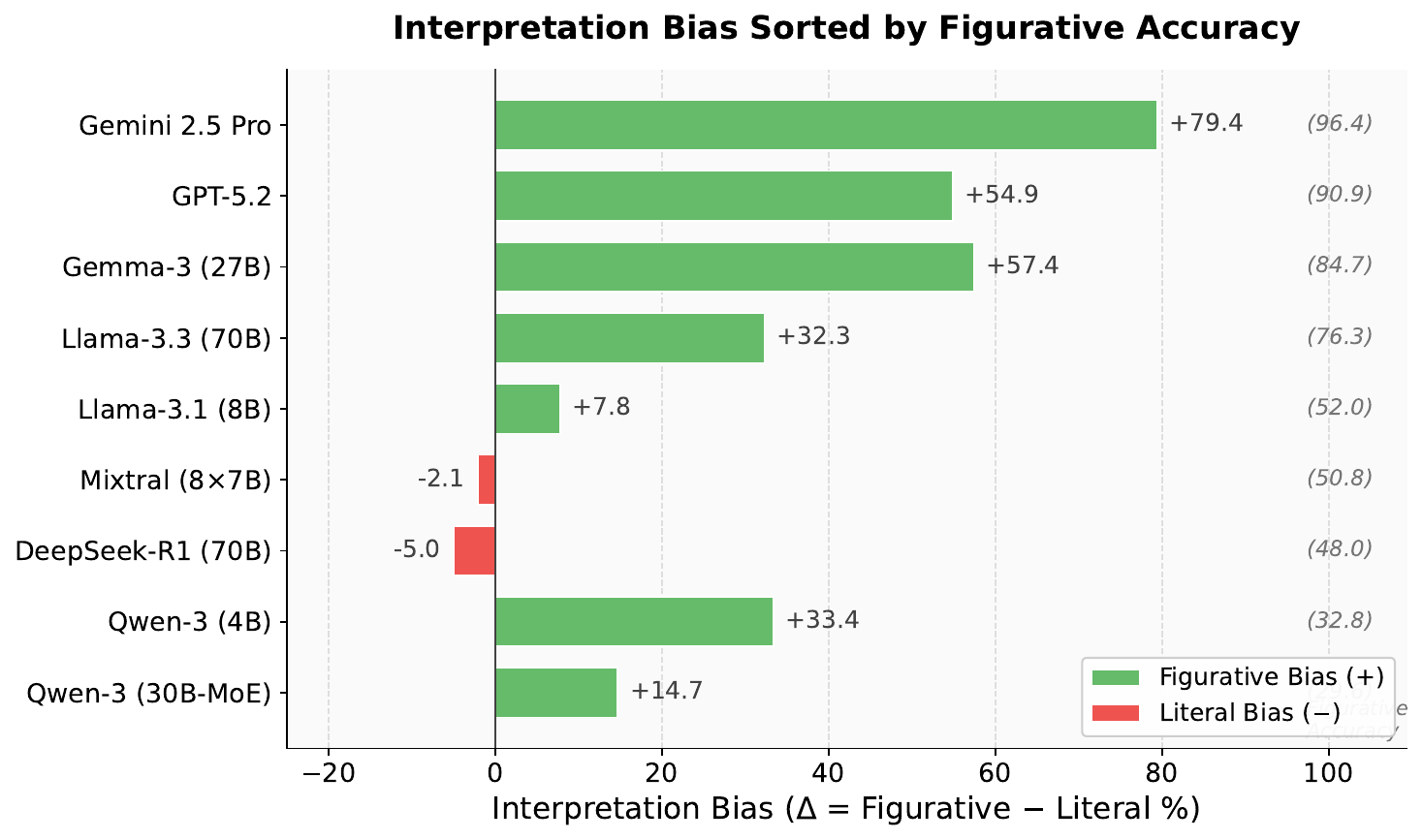}
    \caption{Interpretation bias ($\Delta$) sorted by figurative accuracy. Positive $\Delta$ indicates figurative preference. Scores in parentheses denote weighted accuracy.}
    \label{fig:bias_figurative}
\end{figure}

\begin{figure}[ht]
    \centering
    \includegraphics[width=1\columnwidth]{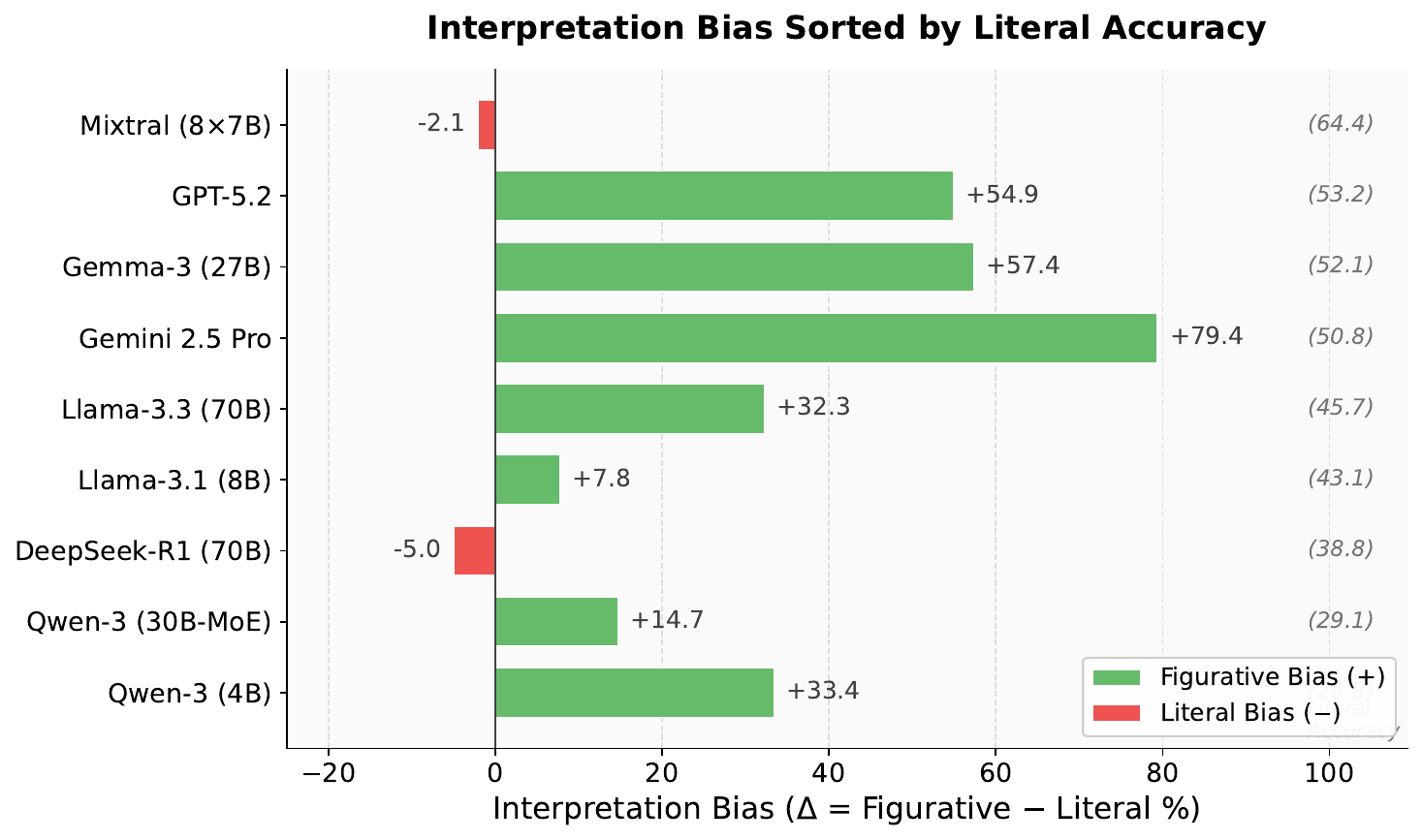}
    \caption{Interpretation bias ($\Delta$) sorted by literal accuracy. Models with lower figurative bias (e.g., Mixtral) tend to perform better on literal interpretation.}
    \label{fig:bias_literal}
\end{figure}

Figures~\ref{fig:bias_figurative} and \ref{fig:bias_literal} visualize interpretation bias sorted by task performance to examine potential correlations. Models with stronger figurative bias generally achieve higher figurative accuracy (Figure~\ref{fig:bias_figurative}), suggesting that default figurative interpretation aligns with task demands. Notably, Mixtral achieves the highest literal accuracy despite near-neutral bias, indicating that balanced interpretation may benefit literal comprehension (Figure~\ref{fig:bias_literal}).

\section{Activation Steering Details}
\label{appx:steering_details}

\definecolor{svHigh}{HTML}{C8E6C9}   
\definecolor{svMid}{HTML}{FFF9C4}    
\definecolor{svLow}{HTML}{FFCDD2}    
\definecolor{svRow}{HTML}{F5F5F5}    
\definecolor{svPos}{HTML}{008026}    

\providecommand{\dpos}[1]{\textcolor{svPos}{(#1)}}
\providecommand{\dneg}[1]{\textcolor{red}{(#1)}}
\providecommand{\dzero}[1]{\textcolor{gray}{(#1)}}

\begin{table*}[t]
\centering
\small
\setlength{\tabcolsep}{4pt}
\renewcommand{\arraystretch}{1.15}

{\rowcolors{1}{}{}%
\begin{tabular}{@{}lcccccc@{}}
\toprule
& \multicolumn{3}{c}{\textbf{Sentence (Accuracy \%)}} & \multicolumn{3}{c}{\textbf{Conversation (Accuracy \%)}} \\
\cmidrule(lr){2-4} \cmidrule(lr){5-7}
\textbf{Model / Setting}
& \cellcolor{svHigh}\textbf{High}
& \cellcolor{svMid}\textbf{Mid}
& \cellcolor{svLow}\textbf{Low}
& \cellcolor{svHigh}\textbf{High}
& \cellcolor{svMid}\textbf{Mid}
& \cellcolor{svLow}\textbf{Low} \\
\midrule

\rowcolor{white}\multicolumn{7}{@{}l}{\textit{\small Reasoning + Memorization}} \\
Qwen-3 (4B) & 61.86 & 71.82 & 44.90 & 67.84 & 59.66 & 38.31 \\
\rowcolor{svRow}
\quad + memorization vect
& 62.89 \dpos{+1.03} & 72.49 \dpos{+0.67} & 46.24 \dpos{+1.34}
& \textbf{68.63} \dpos{+0.79} & 59.83 \dpos{+0.17} & 40.04 \dpos{+1.73} \\
\quad + reasoning vect
& 62.82 \dpos{+0.96} & 72.49 \dpos{+0.67} & 46.44 \dpos{+1.54}
& 68.55 \dpos{+0.71} & 60.33 \dpos{+0.67} & 39.96 \dpos{+1.65} \\
\addlinespace[2pt]

Llama-3.1 (8B) & 63.48 & 78.63 & 52.84 & 62.17 & 77.59 & 51.77 \\
\rowcolor{svRow}
\quad + memorization vect
& \textbf{63.97} \dpos{+0.49} & \textbf{79.50} \dpos{+0.87} & 53.30 \dpos{+0.46}
& 62.61 \dpos{+0.44} & 77.59 \dzero{+0.00} & 52.22 \dpos{+0.45} \\
\quad + reasoning vect
& 63.92 \dpos{+0.44} & \textbf{79.50} \dpos{+0.87} & \textbf{53.36} \dpos{+0.52}
& 62.54 \dpos{+0.37} & \textbf{77.75} \dpos{+0.16} & \textbf{52.37} \dpos{+0.60} \\

\hdashline
\noalign{\vskip 2pt}
\rowcolor{white}\multicolumn{7}{@{}l}{\textit{\small Reasoning}} \\
Qwen-3 (4B) & 97.93 & \textbf{98.33} & 83.32 & \textbf{98.11} & 98.00 & 83.73 \\
\rowcolor{svRow}
\quad + memorization vect
& 97.93 \dzero{+0.00} & \textbf{98.33} \dzero{+0.00} & 83.90 \dpos{+0.58}
& \textbf{98.11} \dzero{+0.00} & \textbf{98.67} \dpos{+0.67} & 84.06 \dpos{+0.33} \\
\quad + reasoning vect
& \textbf{98.04} \dpos{+0.11} & \textbf{98.33} \dzero{+0.00} & 83.55 \dpos{+0.23}
& \textbf{98.11} \dzero{+0.00} & 98.33 \dpos{+0.33} & 84.25 \dpos{+0.51} \\
\addlinespace[2pt]

Llama-3.1 (8B) & 97.62 & 97.69 & 87.29 & 96.00 & 97.36 & 86.05 \\
\rowcolor{svRow}
\quad + memorization vect
& 97.62 \dzero{+0.00} & 98.02 \dpos{+0.33} & 87.41 \dpos{+0.12}
& 96.00 \dzero{+0.00} & 97.36 \dzero{+0.00} & \textbf{86.42} \dpos{+0.38} \\
\quad + reasoning vect
& 97.62 \dzero{+0.00} & 97.69 \dzero{+0.00} & \textbf{87.51} \dpos{+0.22}
& 96.11 \dpos{+0.11} & 97.36 \dzero{+0.00} & 86.17 \dpos{+0.13} \\

\hdashline
\noalign{\vskip 2pt}
\rowcolor{white}\multicolumn{7}{@{}l}{\textit{\small Memorization}} \\
Qwen-3 (4B) & \multicolumn{3}{c}{\textemdash} & 86.62 & 66.22 & 43.37 \\
\rowcolor{svRow}
\quad + memorization vect
& \multicolumn{3}{c}{\textemdash} & \textbf{88.03} \dpos{+1.41} & 66.53 \dpos{+0.31} & 45.00 \dpos{+1.64} \\
\quad + reasoning vect
& \multicolumn{3}{c}{\textemdash} & 87.58 \dpos{+0.74}  & \textbf{66.86} \dpos{+0.64} & 44.80 \dpos{+1.35}\\
\addlinespace[2pt]

Llama-3.1 (8B) & \multicolumn{3}{c}{\textemdash} & 85.34 & 62.25 & 44.14 \\
\rowcolor{svRow}
\quad + memorization vect
& \multicolumn{3}{c}{\textemdash} & 85.45 \dpos{+0.11} & 62.56 \dpos{+0.31} & \textbf{45.62} \dpos{+1.48} \\
\quad + reasoning vect
& \multicolumn{3}{c}{\textemdash} & 85.34 \dzero{+0.00} & 63.20 \dpos{+0.95} & 45.37 \dpos{+1.23} \\

\bottomrule
\end{tabular}%
}

\caption{\textbf{MMLU-Pro steering vectors} applied to to both conversation and sentence tasks. Values are language accuracies (\%) aggregated over high-/mid-/low-resource languages under sentence and conversation evaluation. Parentheses denote absolute change vs.\ the unsteered baseline for the same model and split. \textbf{Bold} marks the best score across both models (within each evaluation split and column).}
\label{tab:mmlu_pro_steering_vectors}
\end{table*}

\definecolor{svHigh}{HTML}{C8E6C9}   
\definecolor{svMid}{HTML}{FFF9C4}    
\definecolor{svLow}{HTML}{FFCDD2}    
\definecolor{svRow}{HTML}{F5F5F5}    
\definecolor{svPos}{HTML}{008026}    

\providecommand{\dpos}[1]{\textcolor{svPos}{(#1)}}
\providecommand{\dneg}[1]{\textcolor{red}{(#1)}}
\providecommand{\dzero}[1]{\textcolor{gray}{(#1)}}

\begin{table*}[t]
\centering
\small
\setlength{\tabcolsep}{4pt}
\renewcommand{\arraystretch}{1.15}

{\rowcolors{1}{}{}%
\begin{tabular}{@{}lcccccc@{}}
\toprule
& \multicolumn{3}{c}{\textbf{Sentence (Accuracy \%)}} & \multicolumn{3}{c}{\textbf{Conversation (Accuracy \%)}} \\
\cmidrule(lr){2-4} \cmidrule(lr){5-7}
\textbf{Model / Setting}
& \cellcolor{svHigh}\textbf{High}
& \cellcolor{svMid}\textbf{Mid}
& \cellcolor{svLow}\textbf{Low}
& \cellcolor{svHigh}\textbf{High}
& \cellcolor{svMid}\textbf{Mid}
& \cellcolor{svLow}\textbf{Low} \\
\midrule

\rowcolor{white}\multicolumn{7}{@{}l}{\textit{\small Reasoning + Memorization}} \\
Qwen-3 (4B) & 61.86 & 71.82 & 44.90 & 67.84 & 59.66 & 38.31 \\
\rowcolor{svRow}
\quad + memorization vect
& 62.56 \dpos{+0.70} & 72.49 \dpos{+0.67} & 46.32 \dpos{+1.42}
& \textbf{68.67} \dpos{+0.83} & 60.00 \dpos{+0.34} & 39.51 \dpos{+1.19} \\
\quad + reasoning vect
& 62.83 \dpos{+0.97} & 72.20 \dpos{+0.38} & 46.25 \dpos{+1.35}
& 68.43 \dpos{+0.59} & 60.00 \dpos{+0.34} & 39.26 \dpos{+0.95} \\
\addlinespace[2pt]

Llama-3.1 (8B) & 63.48 & 78.63 & 52.84 & 62.17 & 77.59 & 51.77 \\
\rowcolor{svRow}
\quad + memorization vect
& \textbf{63.97} \dpos{+0.49} & \textbf{79.50} \dpos{+0.87} & \textbf{53.36} \dpos{+0.52}
& 62.83 \dpos{+0.66} & \textbf{77.75} \dpos{+0.16} & \textbf{52.38} \dpos{+0.61} \\
\quad + reasoning vect
& 63.82 \dpos{+0.34} & 79.33 \dpos{+0.70} & 53.28 \dpos{+0.44}
& 62.64 \dpos{+0.47} & \textbf{77.75} \dpos{+0.16} & 52.33 \dpos{+0.56} \\

\hdashline
\noalign{\vskip 2pt}
\rowcolor{white}\multicolumn{7}{@{}l}{\textit{\small Reasoning}} \\
Qwen-3 (4B) & 97.93 & \textbf{98.33} & 83.32 & \textbf{98.11} & 98.00 & 83.73 \\
\rowcolor{svRow}
\quad + memorization vect
& 97.93 \dzero{+0.00} & \textbf{98.33} \dzero{+0.00} & 83.90 \dpos{+0.58}
& 98.00 \dneg{-0.11} & \textbf{98.33} \dpos{+0.33} & 84.12 \dpos{+0.39} \\
\quad + reasoning vect
& \textbf{98.04} \dpos{+0.11} & \textbf{98.33} \dzero{+0.00} & 83.55 \dpos{+0.23}
& \textbf{98.11} \dzero{+0.00} & \textbf{98.33} \dpos{+0.33} & 84.12 \dpos{+0.39} \\
\addlinespace[2pt]

Llama-3.1 (8B) & 97.62 & 97.69 & 87.29 & 96.00 & 97.36 & 86.05 \\
\rowcolor{svRow}
\quad + memorization vect
& \textbf{97.62} \dzero{+0.00} & 98.02 \dpos{+0.33} & 87.41 \dpos{+0.12}
& 96.11 \dpos{+0.11} & 97.36 \dzero{+0.00} & 86.07 \dpos{+0.03} \\
\quad + reasoning vect
& \textbf{97.62} \dzero{+0.00} & 97.69 \dzero{+0.00} & \textbf{87.51} \dpos{+0.22}
& 96.11 \dpos{+0.11} & 97.36 \dzero{+0.00} & \textbf{86.17} \dpos{+0.13} \\

\hdashline
\noalign{\vskip 2pt}
\rowcolor{white}\multicolumn{7}{@{}l}{\textit{\small Memorization}} \\
Qwen-3 (4B) & \multicolumn{3}{c}{\textemdash} & 86.62 & 66.22 & 43.37 \\
\rowcolor{svRow}
\quad + memorization vect
& \multicolumn{3}{c}{\textemdash} & \textbf{87.58} \dpos{+0.96} & \textbf{66.53} \dpos{+0.31} & 44.53 \dpos{+1.16} \\
\quad + reasoning vect
& \multicolumn{3}{c}{\textemdash} & 87.32 \dpos{+0.69} & \textbf{66.53} \dpos{+0.31} & 44.78 \dpos{+1.42} \\
\addlinespace[2pt]

Llama-3.1 (8B) & \multicolumn{3}{c}{\textemdash} & 85.34 & 62.25 & 44.14 \\
\rowcolor{svRow}
\quad + memorization vect
& \multicolumn{3}{c}{\textemdash} & 85.34 \dzero{+0.00} & 63.22 \dpos{+0.98} & 45.51 \dpos{+1.37} \\
\quad + reasoning vect
& \multicolumn{3}{c}{\textemdash} & 85.45 \dpos{+0.11} & 62.86 \dpos{+0.62} & \textbf{45.58} \dpos{+1.44} \\

\bottomrule
\end{tabular}%
}

\caption{\textbf{MIDI steering vectors} for separating memorization and reasoning. Values are language accuracies (\%) aggregated over high-/mid-/low-resource languages under sentence and conversation evaluation. Parentheses denote absolute change vs.\ the unsteered baseline for the same model and split. \textbf{Bold} marks the best score across both models (within each evaluation split and column).}
\label{tab:custom_steering_vectors}
\end{table*}


\definecolor{highres}{HTML}{C8E6C9}    
\definecolor{medres}{HTML}{FFF9C4}     
\definecolor{lowres}{HTML}{FFCDD2}     
\definecolor{rowgray}{HTML}{F5F5F5}    
\definecolor{PosGreen}{HTML}{008026}

\renewcommand{\dpos}[1]{\textcolor{PosGreen}{(#1)}}
\renewcommand{\dneg}[1]{\textcolor{red}{(#1)}}
\renewcommand{\dzero}[1]{\textcolor{gray}{(#1)}}

\begin{table*}[t]
\centering
\vspace{-2mm}
\small
\setlength{\tabcolsep}{3pt}
\renewcommand{\arraystretch}{1.12}
\resizebox{0.95\textwidth}{!}{%
\begin{tabular}{@{}l c r r r @{\hspace{10pt}} l c r r r@{}}
\toprule
\multicolumn{5}{c}{\textbf{Qwen-3 (4B)}} & \multicolumn{5}{c}{\textbf{Llama-3.1 (8B)}} \\
\cmidrule(lr){1-5}\cmidrule(lr){6-10}
\textbf{Language} & \textbf{Resource class.} & \textbf{Base} & \textbf{+Mem} & \textbf{+Reas} &
\textbf{Language} & \textbf{Resource class.} & \textbf{Base} & \textbf{+Mem} & \textbf{+Reas} \\
\midrule
 Chinese & \cellcolor{highres}\textbf{High} & 68.00 & 68.50 \dpos{+0.50} & 68.50 \dpos{+0.50} & Chinese & \cellcolor{highres}\textbf{High} & 59.17 & 59.33 \dpos{+0.16} & 59.83 \dpos{+0.66} \\
\rowcolor{rowgray}  Japanese & \cellcolor{highres}\textbf{High} & 71.88 & 72.32 \dpos{+0.44} & 72.77 \dpos{+0.89} & Japanese & \cellcolor{highres}\textbf{High} & 66.07 & 66.96 \dpos{+0.89} & 66.52 \dpos{+0.45} \\
 Russian & \cellcolor{highres}\textbf{High} & 63.66 & 65.08 \dpos{+1.42} & 64.37 \dpos{+0.71} & Russian & \cellcolor{highres}\textbf{High} & 61.28 & 61.52 \dpos{+0.24} & 61.28 \dzero{+0.00} \\
\rowcolor{rowgray}  Arabic\_UAE & \cellcolor{medres}\textbf{Mid} & 23.23 & 23.74 \dpos{+0.51} & 23.23 \dzero{+0.00} & Arabic\_UAE & \cellcolor{medres}\textbf{Mid} & 60.61 & 60.61 \dzero{+0.00} & 61.11 \dpos{+0.50} \\
 Indonesia & \cellcolor{medres}\textbf{Mid} & 75.76 & 75.76 \dzero{+0.00} & 75.76 \dzero{+0.00} & Indonesia & \cellcolor{medres}\textbf{Mid} & 75.15 & 75.15 \dzero{+0.00} & 75.15 \dzero{+0.00} \\
\rowcolor{rowgray}  Vietnam & \cellcolor{medres}\textbf{Mid} & 80.00 & 80.00 \dzero{+0.00} & 82.00 \dpos{+2.00} & Vietnam & \cellcolor{medres}\textbf{Mid} & 97.00 & 97.00 \dzero{+0.00} & 97.00 \dzero{+0.00} \\
 Arabic\_Egypt & \cellcolor{lowres}\textbf{Low} & 29.35 & 31.52 \dpos{+2.17} & 31.52 \dpos{+2.17} & Arabic\_Egypt & \cellcolor{lowres}\textbf{Low} & 59.24 & 59.24 \dzero{+0.00} & 59.24 \dzero{+0.00} \\
\rowcolor{rowgray}  Arabic\_Morocco & \cellcolor{lowres}\textbf{Low} & 25.39 & 26.42 \dpos{+1.03} & 26.42 \dpos{+1.03} & Arabic\_Morocco & \cellcolor{lowres}\textbf{Low} & 57.51 & 57.51 \dzero{+0.00} & 57.51 \dzero{+0.00} \\
 Arabic\_Syrian & \cellcolor{lowres}\textbf{Low} & 25.00 & 27.00 \dpos{+2.00} & 26.50 \dpos{+1.50} & Arabic\_Syrian & \cellcolor{lowres}\textbf{Low} & 55.50 & 56.00 \dpos{+0.50} & 56.50 \dpos{+1.00} \\
\rowcolor{rowgray}  Persian & \cellcolor{lowres}\textbf{Low} & 51.09 & 51.63 \dpos{+0.54} & 52.72 \dpos{+1.63} & Persian & \cellcolor{lowres}\textbf{Low} & 65.22 & 65.22 \dzero{+0.00} & 65.76 \dpos{+0.54} \\
 Javanese & \cellcolor{lowres}\textbf{Low} & 43.00 & 45.50 \dpos{+2.50} & 45.50 \dpos{+2.50} & Javanese & \cellcolor{lowres}\textbf{Low} & 54.00 & 54.50 \dpos{+0.50} & 54.50 \dpos{+0.50} \\
\rowcolor{rowgray}  Kannada & \cellcolor{lowres}\textbf{Low} & 61.56 & 63.32 \dpos{+1.76} & 63.82 \dpos{+2.26} & Kannada & \cellcolor{lowres}\textbf{Low} & 67.34 & 67.84 \dpos{+0.50} & 68.34 \dpos{+1.00} \\
 Kazakh & \cellcolor{lowres}\textbf{Low} & 46.50 & 47.50 \dpos{+1.00} & 47.50 \dpos{+1.00} & Kazakh & \cellcolor{lowres}\textbf{Low} & 45.00 & 46.50 \dpos{+1.50} & 46.50 \dpos{+1.50} \\
\rowcolor{rowgray}  Minangkabau & \cellcolor{lowres}\textbf{Low} & 27.52 & 30.87 \dpos{+3.35} & 30.20 \dpos{+2.68} & Minangkabau & \cellcolor{lowres}\textbf{Low} & 33.56 & 34.23 \dpos{+0.67} & 34.23 \dpos{+0.67} \\
 Sundanese & \cellcolor{lowres}\textbf{Low} & 27.91 & 30.23 \dpos{+2.32} & 30.23 \dpos{+2.32} & Sundanese & \cellcolor{lowres}\textbf{Low} & 45.93 & 45.93 \dzero{+0.00} & 46.51 \dpos{+0.58} \\
\rowcolor{rowgray}  Tamil & \cellcolor{lowres}\textbf{Low} & 58.67 & 60.20 \dpos{+1.53} & 59.18 \dpos{+0.51} & Tamil & \cellcolor{lowres}\textbf{Low} & 60.71 & 61.73 \dpos{+1.02} & 61.73 \dpos{+1.02} \\
 Telugu & \cellcolor{lowres}\textbf{Low} & 39.13 & 40.94 \dpos{+1.81} & 40.58 \dpos{+1.45} & Telugu & \cellcolor{lowres}\textbf{Low} & 48.91 & 49.64 \dpos{+0.73} & 49.28 \dpos{+0.37} \\
\rowcolor{rowgray}  Yoruba & \cellcolor{lowres}\textbf{Low} & 24.63 & 25.37 \dpos{+0.74} & 25.37 \dpos{+0.74} & Yoruba & \cellcolor{lowres}\textbf{Low} & 28.36 & 28.36 \dzero{+0.00} & 28.36 \dzero{+0.00} \\
\bottomrule
\end{tabular}%
}
\caption{\textbf{Conversation Vanilla Task.} Per-language (country) results for \textbf{MMLU-Pro vector steering}. We report accuracy (\%) for the baseline model (\textbf{Base}) and after applying memorization (\textbf{+Mem}) or reasoning (\textbf{+Reas}) steering; parentheses denote the absolute change vs.\ baseline. For each model and steering type, the intervention is applied at the \textbf{best-performing layer} (selected on a validation set). \textbf{Resource classification} indicates language resource level (\colorbox{highres}{\textbf{High}}, \colorbox{medres}{\textbf{Mid}}, \colorbox{lowres}{\textbf{Low}}).}
\label{tab:mmlu_lang_vanilla_overall}
\end{table*}

\begin{table*}[t]
\centering
\vspace{-2mm}
\small
\setlength{\tabcolsep}{3pt}
\renewcommand{\arraystretch}{1.12}
\resizebox{0.95\textwidth}{!}{%
\begin{tabular}{@{}l c r r r @{\hspace{10pt}} l c r r r@{}}
\toprule
\multicolumn{5}{c}{\textbf{Qwen-3 (4B)}} & \multicolumn{5}{c}{\textbf{Llama-3.1 (8B)}} \\
\cmidrule(lr){1-5}\cmidrule(lr){6-10}
\textbf{Language} & \textbf{Resource class.} & \textbf{Base} & \textbf{+Mem} & \textbf{+Reas} &
\textbf{Language} & \textbf{Resource class.} & \textbf{Base} & \textbf{+Mem} & \textbf{+Reas} \\
\midrule
 Chinese & \cellcolor{highres}\textbf{High} & 95.67 & 95.67 \dzero{+0.00} & 95.67 \dzero{+0.00} & Chinese & \cellcolor{highres}\textbf{High} & 90.67 & 90.67 \dzero{+0.00} & 91.00 \dpos{+0.33} \\
\rowcolor{rowgray}  Japanese & \cellcolor{highres}\textbf{High} & 99.13 & 99.13 \dzero{+0.00} & 99.13 \dzero{+0.00} & Japanese & \cellcolor{highres}\textbf{High} & 98.26 & 98.26 \dzero{+0.00} & 98.26 \dzero{+0.00} \\
 Russian & \cellcolor{highres}\textbf{High} & 99.53 & 99.53 \dzero{+0.00} & 99.53 \dzero{+0.00} & Russian & \cellcolor{highres}\textbf{High} & 99.06 & 99.06 \dzero{+0.00} & 99.06 \dzero{+0.00} \\
\rowcolor{rowgray}  Arabic\_UAE & \cellcolor{medres}\textbf{Mid} & 96.00 & 97.00 \dpos{+1.00} & 96.00 \dzero{+0.00} & Arabic\_UAE & \cellcolor{medres}\textbf{Mid} & 97.00 & 97.00 \dzero{+0.00} & 97.00 \dzero{+0.00} \\
 Indonesia & \cellcolor{medres}\textbf{Mid} & 100.00 & 100.00 \dzero{+0.00} & 100.00 \dzero{+0.00} & Indonesia & \cellcolor{medres}\textbf{Mid} & 99.07 & 99.07 \dzero{+0.00} & 99.07 \dzero{+0.00} \\
\rowcolor{rowgray}  Vietnam & \cellcolor{medres}\textbf{Mid} & 98.00 & 99.00 \dpos{+1.00} & 99.00 \dpos{+1.00} & Vietnam & \cellcolor{medres}\textbf{Mid} & 96.00 & 96.00 \dzero{+0.00} & 96.00 \dzero{+0.00} \\
 Arabic\_Egypt & \cellcolor{lowres}\textbf{Low} & 98.00 & 99.00 \dpos{+1.00} & 99.00 \dpos{+1.00} & Arabic\_Egypt & \cellcolor{lowres}\textbf{Low} & 98.00 & 98.00 \dzero{+0.00} & 98.00 \dzero{+0.00} \\
\rowcolor{rowgray}  Arabic\_Morocco & \cellcolor{lowres}\textbf{Low} & 95.92 & 95.92 \dzero{+0.00} & 95.92 \dzero{+0.00} & Arabic\_Morocco & \cellcolor{lowres}\textbf{Low} & 94.90 & 95.92 \dpos{+1.02} & 95.92 \dpos{+1.02} \\
 Arabic\_Syrian & \cellcolor{lowres}\textbf{Low} & 99.00 & 99.00 \dzero{+0.00} & 99.00 \dzero{+0.00} & Arabic\_Syrian & \cellcolor{lowres}\textbf{Low} & 99.00 & 99.00 \dzero{+0.00} & 99.00 \dzero{+0.00} \\
\rowcolor{rowgray}  Persian & \cellcolor{lowres}\textbf{Low} & 94.12 & 96.08 \dpos{+1.96} & 97.06 \dpos{+2.94} & Persian & \cellcolor{lowres}\textbf{Low} & 98.04 & 98.04 \dzero{+0.00} & 98.04 \dzero{+0.00} \\
 Javanese & \cellcolor{lowres}\textbf{Low} & 99.04 & 99.04 \dzero{+0.00} & 99.04 \dzero{+0.00} & Javanese & \cellcolor{lowres}\textbf{Low} & 97.12 & 97.12 \dzero{+0.00} & 97.12 \dzero{+0.00} \\
\rowcolor{rowgray}  Kannada & \cellcolor{lowres}\textbf{Low} & 96.46 & 96.46 \dzero{+0.00} & 95.96 \dneg{-0.50} & Kannada & \cellcolor{lowres}\textbf{Low} & 98.99 & 99.49 \dpos{+0.50} & 99.49 \dpos{+0.50} \\
 Kazakh & \cellcolor{lowres}\textbf{Low} & 85.00 & 85.00 \dzero{+0.00} & 87.00 \dpos{+2.00} & Kazakh & \cellcolor{lowres}\textbf{Low} & 98.00 & 98.00 \dzero{+0.00} & 98.00 \dzero{+0.00} \\
\rowcolor{rowgray}  Minangkabau & \cellcolor{lowres}\textbf{Low} & 70.00 & 70.00 \dzero{+0.00} & 69.00 \dneg{-1.00} & Minangkabau & \cellcolor{lowres}\textbf{Low} & 68.00 & 68.00 \dzero{+0.00} & 68.00 \dzero{+0.00} \\
 Sundanese & \cellcolor{lowres}\textbf{Low} & 74.00 & 74.00 \dzero{+0.00} & 74.00 \dzero{+0.00} & Sundanese & \cellcolor{lowres}\textbf{Low} & 70.00 & 71.00 \dpos{+1.00} & 70.00 \dzero{+0.00} \\
\rowcolor{rowgray}  Tamil & \cellcolor{lowres}\textbf{Low} & 95.96 & 95.96 \dzero{+0.00} & 95.96 \dzero{+0.00} & Tamil & \cellcolor{lowres}\textbf{Low} & 96.97 & 97.98 \dpos{+1.01} & 96.97 \dzero{+0.00} \\
 Telugu & \cellcolor{lowres}\textbf{Low} & 83.45 & 83.45 \dzero{+0.00} & 84.17 \dpos{+0.72} & Telugu & \cellcolor{lowres}\textbf{Low} & 79.86 & 79.86 \dzero{+0.00} & 79.86 \dzero{+0.00} \\
\rowcolor{rowgray}  Yoruba & \cellcolor{lowres}\textbf{Low} & 13.86 & 14.85 \dpos{+0.99} & 14.85 \dpos{+0.99} & Yoruba & \cellcolor{lowres}\textbf{Low} & 33.66 & 34.65 \dpos{+0.99} & 33.66 \dzero{+0.00} \\
\bottomrule
\end{tabular}%
}
\caption{\textbf{Reasoning task (Conversation).} Per-language (country) results with \textbf{MMLU-Pro vector steering} (same reporting as Table~\ref{tab:mmlu_lang_vanilla_overall}). Parentheses denote the absolute change vs.\ baseline; steering is applied at the \textbf{best-performing layer} selected on a validation set. \textbf{Resource classification} indicates language resource level (\colorbox{highres}{\textbf{High}}, \colorbox{medres}{\textbf{Mid}}, \colorbox{lowres}{\textbf{Low}}).}
\label{tab:mmlu_lang_reasoning_figurative}
\end{table*}

\begin{table*}[t]
\centering
\vspace{-2mm}
\small
\setlength{\tabcolsep}{3pt}
\renewcommand{\arraystretch}{1.12}
\resizebox{0.95\textwidth}{!}{%
\begin{tabular}{@{}l c r r r @{\hspace{10pt}} l c r r r@{}}
\toprule
\multicolumn{5}{c}{\textbf{Qwen-3 (4B)}} & \multicolumn{5}{c}{\textbf{Llama-3.1 (8B)}} \\
\cmidrule(lr){1-5}\cmidrule(lr){6-10}
\textbf{Language} & \textbf{Resource class.} & \textbf{Base} & \textbf{+Mem} & \textbf{+Reas} &
\textbf{Language} & \textbf{Resource class.} & \textbf{Base} & \textbf{+Mem} & \textbf{+Reas} \\
\midrule
 Chinese & \cellcolor{highres}\textbf{High} & 99.33 & 100.00 \dpos{+0.67} & 100.00 \dpos{+0.67} & Chinese & \cellcolor{highres}\textbf{High} & 97.00 & 97.33 \dpos{+0.33} & 97.00 \dzero{+0.00} \\
\rowcolor{rowgray}  Japanese & \cellcolor{highres}\textbf{High} & 86.96 & 89.57 \dpos{+2.61} & 88.70 \dpos{+1.74} & Japanese & \cellcolor{highres}\textbf{High} & 82.61 & 82.61 \dzero{+0.00} & 82.61 \dzero{+0.00} \\
 Russian & \cellcolor{highres}\textbf{High} & 73.58 & 74.53 \dpos{+0.95} & 74.06 \dpos{+0.48} & Russian & \cellcolor{highres}\textbf{High} & 76.42 & 76.42 \dzero{+0.00} & 76.42 \dzero{+0.00} \\
\rowcolor{rowgray}  Arabic\_UAE & \cellcolor{medres}\textbf{Mid} & 53.00 & 53.00 \dzero{+0.00} & 53.00 \dzero{+0.00} & Arabic\_UAE & \cellcolor{medres}\textbf{Mid} & 54.00 & 54.00 \dzero{+0.00} & 55.00 \dpos{+1.00} \\
 Indonesia & \cellcolor{medres}\textbf{Mid} & 66.67 & 67.59 \dpos{+0.92} & 67.59 \dpos{+0.92} & Indonesia & \cellcolor{medres}\textbf{Mid} & 65.74 & 66.67 \dpos{+0.93} & 67.59 \dpos{+1.85} \\
\rowcolor{rowgray}  Vietnam & \cellcolor{medres}\textbf{Mid} & 79.00 & 79.00 \dzero{+0.00} & 80.00 \dpos{+1.00} & Vietnam & \cellcolor{medres}\textbf{Mid} & 67.00 & 67.00 \dzero{+0.00} & 67.00 \dzero{+0.00} \\
 Arabic\_Egypt & \cellcolor{lowres}\textbf{Low} & 57.00 & 59.00 \dpos{+2.00} & 59.00 \dpos{+2.00} & Arabic\_Egypt & \cellcolor{lowres}\textbf{Low} & 68.00 & 70.00 \dpos{+2.00} & 70.00 \dpos{+2.00} \\
\rowcolor{rowgray}  Arabic\_Morocco & \cellcolor{lowres}\textbf{Low} & 59.79 & 60.82 \dpos{+1.03} & 61.86 \dpos{+2.07} & Arabic\_Morocco & \cellcolor{lowres}\textbf{Low} & 57.73 & 59.79 \dpos{+2.06} & 59.79 \dpos{+2.06} \\
 Arabic\_Syrian & \cellcolor{lowres}\textbf{Low} & 54.00 & 53.00 \dneg{-1.00} & 53.00 \dneg{-1.00} & Arabic\_Syrian & \cellcolor{lowres}\textbf{Low} & 51.00 & 52.00 \dpos{+1.00} & 52.00 \dpos{+1.00} \\
\rowcolor{rowgray}  Persian & \cellcolor{lowres}\textbf{Low} & 44.12 & 46.08 \dpos{+1.96} & 44.12 \dzero{+0.00} & Persian & \cellcolor{lowres}\textbf{Low} & 50.98 & 51.96 \dpos{+0.98} & 51.96 \dpos{+0.98} \\
 Javanese & \cellcolor{lowres}\textbf{Low} & 36.54 & 36.54 \dzero{+0.00} & 36.54 \dzero{+0.00} & Javanese & \cellcolor{lowres}\textbf{Low} & 32.69 & 33.65 \dpos{+0.96} & 32.69 \dzero{+0.00} \\
\rowcolor{rowgray}  Kannada & \cellcolor{lowres}\textbf{Low} & 41.41 & 41.92 \dpos{+0.51} & 42.42 \dpos{+1.01} & Kannada & \cellcolor{lowres}\textbf{Low} & 44.95 & 45.96 \dpos{+1.01} & 44.95 \dzero{+0.00} \\
 Kazakh & \cellcolor{lowres}\textbf{Low} & 46.00 & 50.00 \dpos{+4.00} & 50.00 \dpos{+4.00} & Kazakh & \cellcolor{lowres}\textbf{Low} & 45.00 & 46.00 \dpos{+1.00} & 45.00 \dzero{+0.00} \\
\rowcolor{rowgray}  Minangkabau & \cellcolor{lowres}\textbf{Low} & 34.00 & 37.00 \dpos{+3.00} & 37.00 \dpos{+3.00} & Minangkabau & \cellcolor{lowres}\textbf{Low} & 29.00 & 34.00 \dpos{+5.00} & 34.00 \dpos{+5.00} \\
 Sundanese & \cellcolor{lowres}\textbf{Low} & 26.00 & 26.00 \dzero{+0.00} & 26.00 \dzero{+0.00} & Sundanese & \cellcolor{lowres}\textbf{Low} & 30.00 & 32.00 \dpos{+2.00} & 31.00 \dpos{+1.00} \\
\rowcolor{rowgray}  Tamil & \cellcolor{lowres}\textbf{Low} & 53.06 & 55.10 \dpos{+2.04} & 53.06 \dzero{+0.00} & Tamil & \cellcolor{lowres}\textbf{Low} & 43.88 & 44.90 \dpos{+1.02} & 44.90 \dpos{+1.02} \\
 Telugu & \cellcolor{lowres}\textbf{Low} & 41.73 & 43.88 \dpos{+2.15} & 43.88 \dpos{+2.15} & Telugu & \cellcolor{lowres}\textbf{Low} & 46.76 & 47.48 \dpos{+0.72} & 47.48 \dpos{+0.72} \\
\rowcolor{rowgray}  Yoruba & \cellcolor{lowres}\textbf{Low} & 26.73 & 30.69 \dpos{+3.96} & 28.71 \dpos{+1.98} & Yoruba & \cellcolor{lowres}\textbf{Low} & 29.70 & 29.70 \dzero{+0.00} & 30.69 \dpos{+0.99} \\
\bottomrule
\end{tabular}%
}
\caption{\textbf{Memorization task.} Per-language (country) results with \textbf{MMLU-Pro vector steering} (same reporting as Table~\ref{tab:mmlu_lang_vanilla_overall}). Parentheses denote the absolute change vs.\ baseline; steering is applied at the \textbf{best-performing layer} selected on a validation set. \textbf{Resource classification} indicates language resource level (\colorbox{highres}{\textbf{High}}, \colorbox{medres}{\textbf{Mid}}, \colorbox{lowres}{\textbf{Low}}).}
\label{tab:mmlu_lang_mem_overall}
\end{table*}

This appendix provides the full steering diagnostics referenced in Section~\ref{sec:steering}: per-task $\Delta$-accuracy grids, layer-sweep statistics, direction magnitudes, and flip-rate analysis.

\subsection{Additional Steering Definitions}
For auxiliary analyses we quantify how well an input aligns with the memorization$\rightarrow$reasoning direction using a normalized projection score~\citep{hong2025reasoningmemorization}:
\begin{equation}
    \hat{r}^{(\ell)} = \frac{r^{(\ell)}}{\|r^{(\ell)}\|_2}, \quad s^{(\ell)}(x)=\hat{r}^{(\ell)\top}h^{(\ell)}(x).
    \label{eq:proj_score}
\end{equation}
We also define an \emph{ablation} operator that removes activation components along the reasoning direction (not used in our primary experiments):
\begin{equation}
    h^{(\ell)}(x) \leftarrow h^{(\ell)}(x) - \hat{r}^{(\ell)}\left(\hat{r}^{(\ell)\top}h^{(\ell)}(x)\right).
    \label{eq:ablation}
\end{equation}

\subsection{Where Steering Helps}
\begin{figure*}[t]
    \centering
    \includegraphics[width=\textwidth]{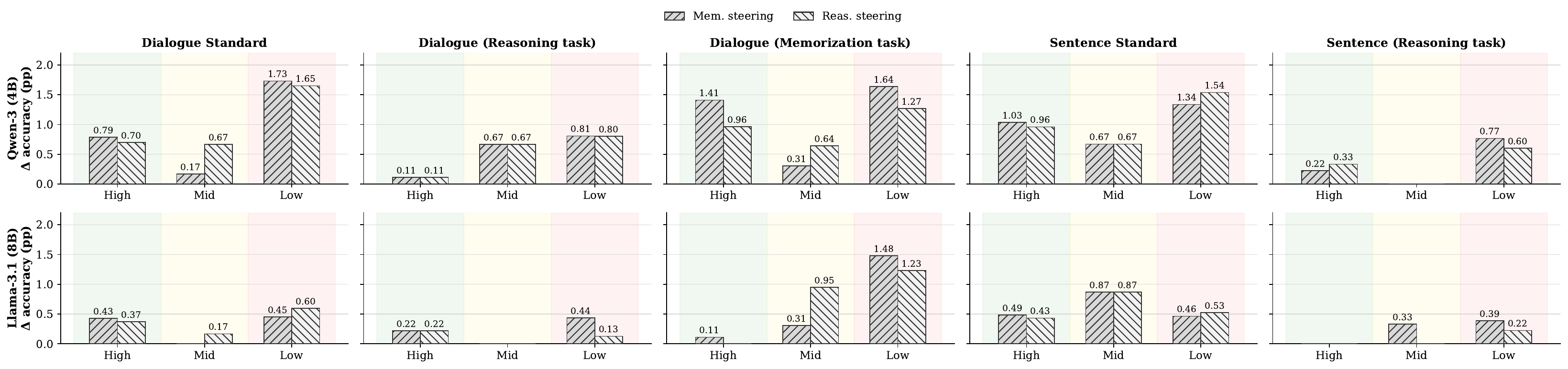}
    \caption{Steering-induced accuracy gains ($\Delta$) using \textbf{MMLU-Pro vectors}, broken down by task and resource tier.}
    \label{fig:delta_grid_mmlu}
\end{figure*}

\begin{figure*}[t]
    \centering
    \includegraphics[width=\textwidth]{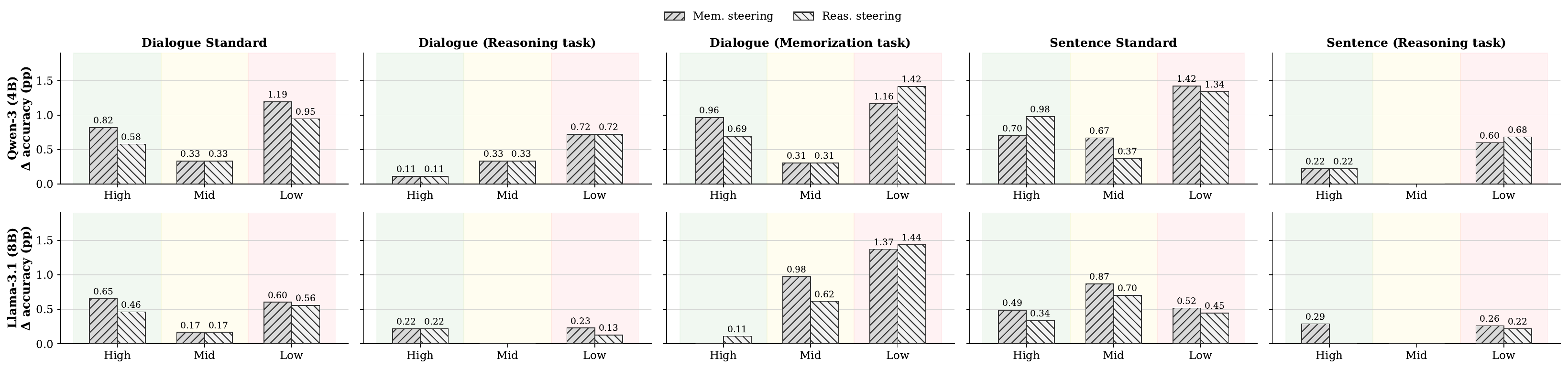}
    \caption{Steering-induced accuracy gains ($\Delta$) using \textbf{MIDI-derived vectors}, broken down by task and resource tier.}
    \label{fig:delta_grid_custom}
\end{figure*}

Figures~\ref{fig:delta_grid_mmlu} and~\ref{fig:delta_grid_custom} report steering-induced accuracy changes (in percentage points), stratified by resource tier.
A consistent pattern emerges across tasks: \textbf{low-resource languages exhibit the largest average gains}, while high-resource languages show more modest changes.
This aligns with a natural \emph{headroom} effect, as high-resource settings often approach ceiling performance, leaving limited room for improvement.

\subsection{Layer Sensitivity and Direction Magnitude}
\begin{figure*}[t]
    \centering
    \includegraphics[width=\textwidth]{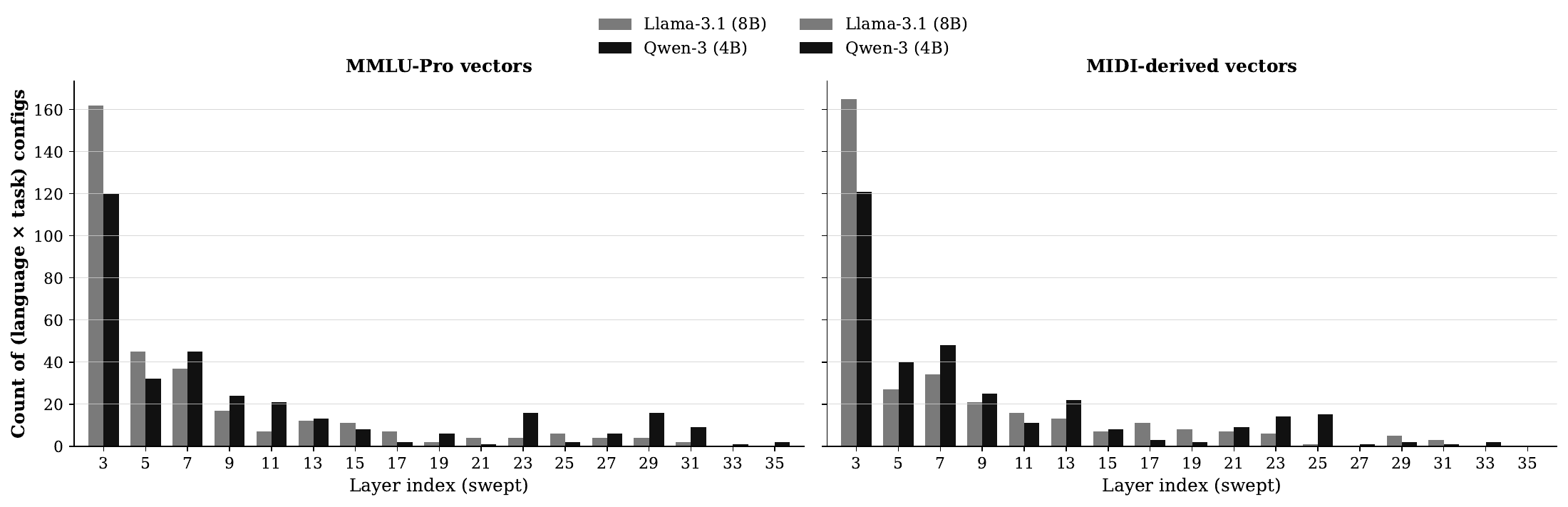}
    \caption{Distribution of best-performing layers (selected by highest development accuracy) from the layer sweep. We evaluate every other layer starting at $\ell{=}3$ (Llama: through $\ell{=}31$; Qwen: through $\ell{=}35$).}
    \label{fig:best_layer_hist}
\end{figure*}

Steering effectiveness is \textbf{highly layer-dependent}: optimal layers vary across languages and tasks.
Figure~\ref{fig:best_layer_hist} shows the distribution of best-performing layers from our sweep.
Within each model, the best-layer distributions for \textbf{MMLU-Pro} and \textbf{MIDI-derived} vectors are strikingly similar, supporting the hypothesis that both probe a shared internal component related to the memorization--reasoning tradeoff (rather than dataset-specific artifacts).

For both models, layer~3 emerges as the most frequently optimal choice.
Median best layers cluster at 3--4 for Llama-3.1~8B (median of 4 with MMLU-Pro vectors; median of 3 with MIDI-derived vectors), whereas Qwen-3~4B exhibits a median of 7 for both vector sources, reflecting a broader distribution toward mid-depth layers.

Notably, suboptimal layer choices can be actively harmful.
Across all language and task configurations, the \emph{worst} layer in each sweep reduces accuracy by $-0.88$ points on average, with the most severe case yielding a $-6.67$ point drop.
This underscores that steering should be approached as a calibrated intervention rather than assumed to yield unconditional improvements.

\begin{figure*}[t]
    \centering
    \includegraphics[width=\textwidth]{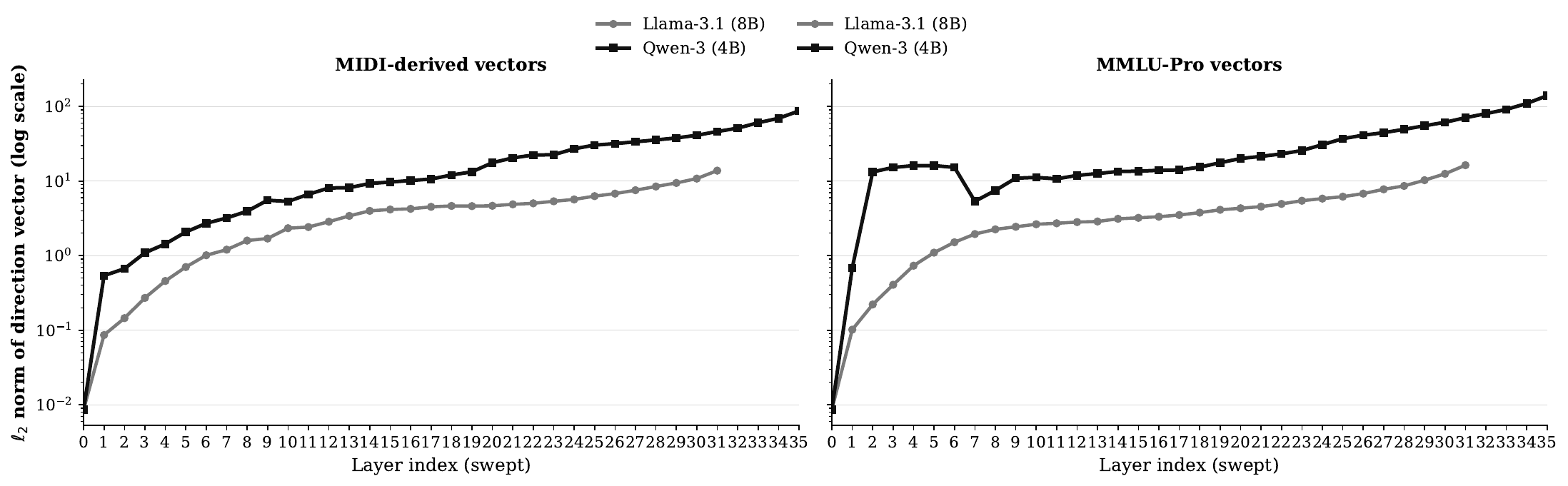}
    \caption{$\ell_2$ norms of the memorization$\rightarrow$reasoning direction vector across swept layers (log scale).}
    \label{fig:direction_norms}
\end{figure*}

Figure~\ref{fig:direction_norms} plots the $\ell_2$ norms of direction vectors across layers.
When averaged across layers (excluding layer~0, where some directions are exactly zero), Qwen's direction vectors exhibit substantially larger norms than Llama's: approximately $5.0\times$ larger for MIDI-derived vectors and $7.4\times$ larger for MMLU-Pro vectors.
This disparity is consistent with Qwen's heightened behavioral sensitivity to steering at equivalent scales.

\subsection{Flip-Rate Analysis}
To quantify how frequently steering alters model predictions, we compute \textbf{flip rates} on the complete evaluation set:
the proportion of examples where the steered prediction differs from the baseline.
We disaggregate these into \emph{improvement flips} (baseline incorrect $\rightarrow$ steered correct) and \emph{regression flips} (baseline correct $\rightarrow$ steered incorrect).

\begin{figure*}[t]
    \centering
    \includegraphics[width=\textwidth]{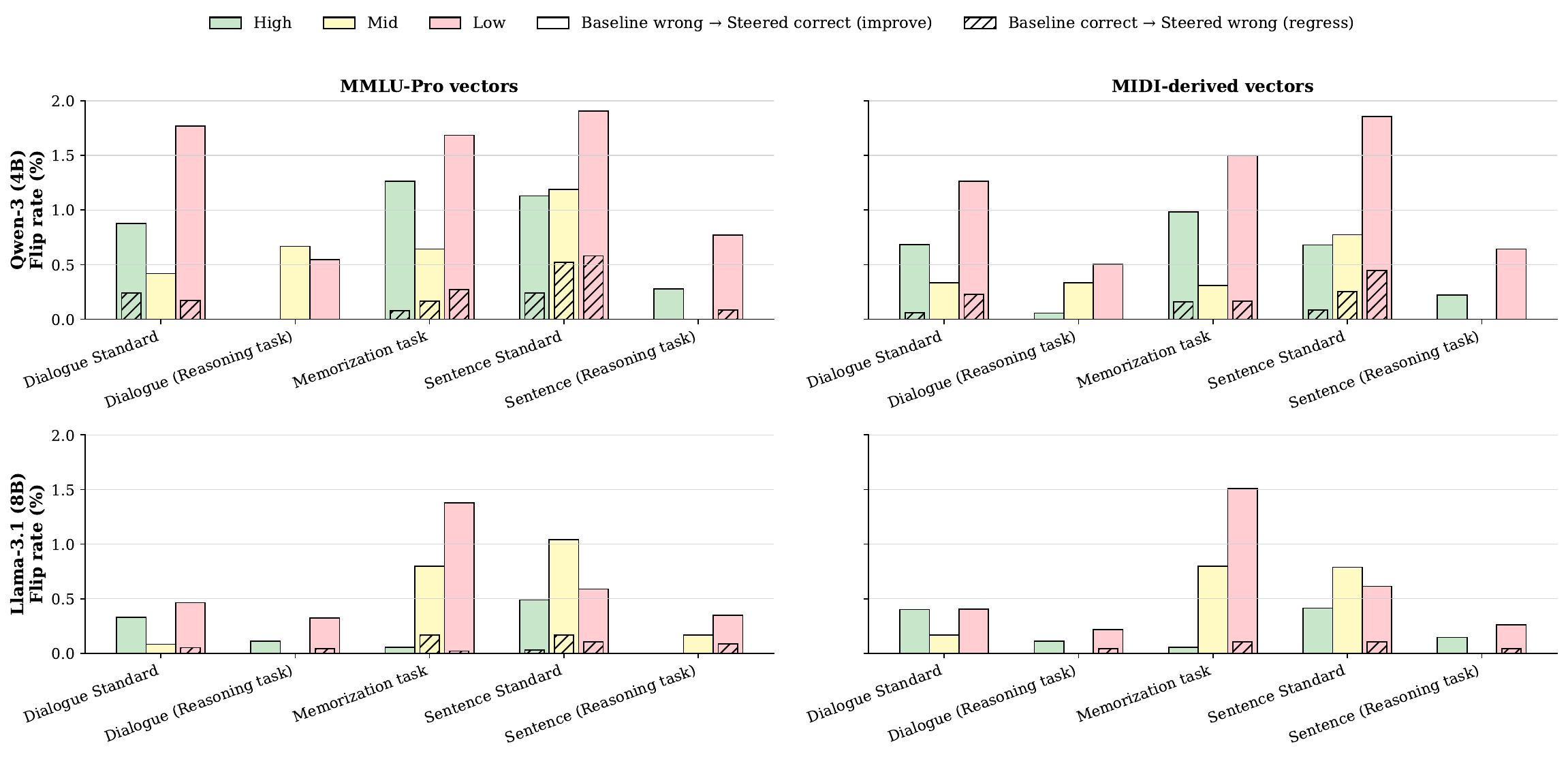}
    \caption{Flip rates on the full evaluation set, macro-averaged across languages within each resource tier and subsequently averaged over tasks. Improvement flips appear as solid bars; regression flips are indicated with hatched overlays.}
    \label{fig:flip_rates_full}
\end{figure*}

Figure~\ref{fig:flip_rates_full} presents flip rates by resource tier, averaged across steering types.
Flip rates remain modest in absolute terms (typically below a few percent), but are systematically elevated in low-resource settings, paralleling the larger accuracy gains in Figures~\ref{fig:delta_grid_mmlu} and~\ref{fig:delta_grid_custom}.

\section{Human Evaluation Details}
\label{app:human_eval}

To verify that \textbf{MIDI}'s idiom comprehension task setup (Section~\ref{sec:main_experiment_mcq}; see format~\ref{app:prompt_default}) is clear and interpretable for human annotators, and to establish a reference point against which model performance can be contextualized, we conducted a human evaluation on a 10\% random sample of idioms drawn from nine languages spanning all three resource tiers: Japanese and Russian (high-resource); Arabic (UAE), Indonesian and Vietnamese (mid-resource); and Arabic (Egypt), Arabic (Morocco), Minangkabau and Yoruba (low-resource). For each language, native speakers answered the same multiple-choice questions used in our model evaluation, covering both sentence- and dialogue-level contexts as well as both figurative and literal idiom usages (with the exception of Vietnamese, which has no literal counterparts in \textbf{MIDI}).

\paragraph{Detailed Human Performance.}
Table~\ref{tab:human_eval} presents the human evaluation results broken down by language. Annotators perform consistently well, with an average overall accuracy of 94\% and per-language accuracies ranging from 88\% (Japanese) to 100\% (Vietnamese). Performance is similarly strong for figurative (97\%) and literal (89\%) usages, as well as across sentence (92\%) and dialogue (96\%) contexts, suggesting that neither the interpretation type of the idiom nor the context leads to systematic ambiguity in the task. The small gap between figurative and literal scores mirrors a trend also seen in some models, where literal counterparts, while grammatically correct, may still prompt a figurative interpretation even among highly proficient speakers.


\paragraph{Detailed Best Models Performance.}
To keep the comparison between humans and models as controlled as possible, we also re-assess the best overall proprietary model, \textit{Gemini 2.5 Pro}, and the best overall open-source model, \textit{Gemma-3 (27B)}, on the exact same 10\% subset that was labeled by human annotators. Tables~\ref{tab:human_eval_closed} and~\ref{tab:human_eval_open} presents the resulting sample-level scores, broken down by context and usage type. On this subset, Gemini 2.5 Pro averages 80\% overall, while Gemma-3 averages 78\%, both well below the 94\% human average. Both models exhibit the same qualitative pattern seen on the full dataset: strong figurative performance (95\% and 90\% respectively) is coupled with a sharp drop on literal usages (both at 59\%), and dialogue context is easier than sentence context (85/76 for Gemini, 84/72 for Gemma-3). The open-source model is most fragile in low-resource languages, falling to 19\% in Yoruba despite Gemini 2.5 Pro retaining 89\% in the same samples.

\paragraph{Human and Model Comparison.}
We contextualize human performance against the top models at two levels of granularity. First, Table~\ref{tab:human_vs_models} compares human accuracy on the 10\% sample with the best-performing proprietary and open-source models for each language, \emph{all evaluated on the same samples}; this yields a strictly matched and aligned comparison. Second, Table~\ref{tab:human_vs_models_full} compares the same human scores in 10\% samples to the corresponding best-model accuracies on the \emph{full} per-language evaluation set from Table~\ref{tab:per_language_accuracy}, and includes a per-row $\Delta$ value that captures the difference between the two gap estimates. Small $\Delta$ values indicate that model accuracy on the 10\% sample closely mirrors performance on the full dataset, serving as a consistency check for the sample-based comparison.

In the 10\% sample comparison, human performance surpasses that of the strongest models in eight out of nine languages, with Vietnamese being the only case of parity (humans 100\%; Gemini 2.5 Pro and Gemma-3 both 100\%). On average, humans score 13 percentage points higher than the best proprietary model and 22 percentage points higher than the best open-source model. The largest human--model discrepancies occur for Russian (+34 relative to the best proprietary model) and Yoruba (+73 relative to the best open-source model), highlighting considerable remaining room for improvement in both high- and low-resource tiers, and in particular for open-source models on low-resource languages.

The full dataset comparison in Table~\ref{tab:human_vs_models_full} is broadly consistent with the 10\% sample findings: most $\Delta$ values lie within about $\pm$5 points, suggesting that the 10\% sample reflects overall dataset behavior reasonably well. A small number of languages exhibit larger deviations in both directions. For Minangkabau (open-source, $\Delta$ $-$14) and Yoruba (proprietary, $\Delta$ $-$9), the MCQs sampled were easier for the model than those in the full dataset. In contrast, for Arabic (Morocco) (proprietary, $\Delta$ +8), the MCQs samples were more difficult. Such deviations are expected given the limited 10\% sample size per language and do not alter the tier-level conclusions.

\paragraph{Interpretation.}
These results support two main conclusions. First, the MCQ task setup is interpretable and can be consistently solved by native speakers across both context formats and usage types, suggesting that model failures on \textbf{MIDI} cannot be explained by task ambiguity or problems with the annotations. Second, current LLMs, including the strongest available proprietary and open-source models still fall substantially short of human performance on idiom comprehension, with most of the gap arising for literal usages and in low-resource languages. Both trends hold whether models are assessed on the matched 10\% sample or on the full per-language dataset. The nine-language evaluation includes half of the 18 languages in \textbf{MIDI} and covers all three resource levels, making it a representative reference point for the dataset in general.

\begin{table*}[t]
\centering
\setlength{\tabcolsep}{5pt}
\setlength{\abovecaptionskip}{3pt}
\setlength{\belowcaptionskip}{0pt}
\renewcommand{\arraystretch}{1.1}
\small
\resizebox{0.85\textwidth}{!}{%
\begin{tabular}{@{}l cc cc c@{}}
\multicolumn{6}{c}{\textit{Human Evaluation on 10\% Random Sample}} \\
\toprule
& \multicolumn{2}{c}{\textbf{Idiom Usage Context}}
& \multicolumn{2}{c}{\textbf{Idiom Usage Type}}
& \\
\cmidrule(lr){2-3} \cmidrule(lr){4-5}
\textbf{Language}
& \textbf{Sent} & \textbf{Conv}
& \textbf{Fig} & \textbf{Lit}
& \textbf{Overall} \\
\midrule
\multicolumn{6}{@{}l}{\cellcolor{highres}\textbf{High-Resource Languages}} \\
\rowcolor{rowgray}
Japanese         &  78 & 100 &  95 &  80 &  88 \\
Russian          &  98 & 100 & 100 &  98 &  99 \\
\midrule
\multicolumn{6}{@{}l}{\cellcolor{medres}\textbf{Mid-Resource Languages}} \\
\rowcolor{rowgray}
Arabic (UAE)     & 100 &  94 &  95 & 100 &  97 \\
Indonesian       &  82 & 100 &  90 &  92 &  91 \\
\rowcolor{rowgray}
Vietnamese       & 100 & 100 & 100 & N/A & 100 \\
\midrule
\multicolumn{6}{@{}l}{\cellcolor{lowres}\textbf{Low-Resource Languages}} \\
\rowcolor{rowgray}
Arabic (Egypt)   &  95 &  88 & 100 &  82 &  92 \\
Arabic (Morocco)   &  90 &  88 & 95 &  83 &  89 \\
\rowcolor{rowgray}
Minangkabau      &  92 & 100 & 100 &  88 &  96 \\
Yoruba      &  92 & 100 & 100 &  86 &  96 \\
\midrule \rowcolor{rowgray}
\textbf{Average}
& \textbf{92} & \textbf{96}
& \textbf{97} & \textbf{89}
& \textbf{94} \\
\bottomrule
\noalign{\vskip 1pt}
\multicolumn{6}{c}{\small\textit{Sent = Sentence \quad\quad Conv = Conversation \quad\quad Fig = Figurative \quad\quad Lit = Literal}} \\
\end{tabular}%
}
\caption{Human accuracy (\%) on a 10\% random sample of idioms from nine languages spanning all three resource tiers. Scores are reported across idiom usage contexts (Sentence, Conversation), idiom usage types (Figurative, Literal), and overall. Vietnamese is marked N/A for the literal column, as it contains no literal counterparts in \textbf{MIDI}.}
\label{tab:human_eval}
\end{table*}
\begin{table*}[t]
\centering
\setlength{\tabcolsep}{5pt}
\setlength{\abovecaptionskip}{3pt}
\setlength{\belowcaptionskip}{0pt}
\renewcommand{\arraystretch}{1.1}
\small
\resizebox{0.85\textwidth}{!}{%
\begin{tabular}{@{}l cc cc c@{}}
\multicolumn{6}{c}{\textit{Best Proprietary Model (Gemini 2.5 Pro) Evaluation on 10\% Random Sample}} \\
\toprule
 & \multicolumn{2}{c}{\textbf{Idiom Usage Context}}
& \multicolumn{2}{c}{\textbf{Idiom Usage Type}}
& \\
\cmidrule(lr){2-3} \cmidrule(lr){4-5}
\textbf{Language}
& \textbf{Sent} & \textbf{Conv}
& \textbf{Fig} & \textbf{Lit}
& \textbf{Overall} \\
\midrule
\multicolumn{6}{@{}l}{\cellcolor{highres}\textbf{High-Resource Languages}} \\
\rowcolor{rowgray}
Japanese         &  61 & 84 & 100 & 40 & 71 \\
Russian          &  69 & 61 & 100 & 30 & 65 \\
\midrule
\multicolumn{6}{@{}l}{\cellcolor{medres}\textbf{Mid-Resource Languages}} \\
\rowcolor{rowgray}
Arabic (UAE)     & 85 & 82 & 95 & 72 & 84 \\
Indonesian       &  88 & 88 & 95 & 75 & 88 \\
\rowcolor{rowgray}
Vietnamese       & 100 & 100 & 100 & N/A & 100 \\
\midrule
\multicolumn{6}{@{}l}{\cellcolor{lowres}\textbf{Low-Resource Languages}} \\
\rowcolor{rowgray}
Arabic (Egypt)   &  79 & 88 & 100 & 65 & 83 \\
Arabic (Morocco)   &  50 & 77 & 95 & 28 & 62 \\
\rowcolor{rowgray}
Minangkabau      &  69 & 93 & 79 & 88 & 82 \\
Yoruba      &  83 & 93 & 95 & 71 & 89 \\
\midrule \rowcolor{rowgray}
\textbf{Average}
& \textbf{76} & \textbf{85}
& \textbf{95} & \textbf{59}
& \textbf{80} \\
\bottomrule
\noalign{\vskip 1pt}
\multicolumn{6}{c}{\small\textit{Sent = Sentence \quad\quad Conv = Conversation \quad\quad Fig = Figurative \quad\quad Lit = Literal}} \\
\end{tabular}%
}
\caption{Best overall proprietary model (Gemini 2.5 Pro) accuracy (\%) on a 10\% random sample of idioms from nine languages spanning all three resource tiers. Scores are reported across idiom usage contexts (Sentence, Conversation), idiom usage types (Figurative, Literal), and overall. Vietnamese is marked N/A for the literal column, as it contains no literal counterparts in \textbf{MIDI}.}
\label{tab:human_eval_closed}
\end{table*}
\begin{table*}[t]
\centering
\setlength{\tabcolsep}{5pt}
\setlength{\abovecaptionskip}{3pt}
\setlength{\belowcaptionskip}{0pt}
\renewcommand{\arraystretch}{1.1}
\small
\resizebox{0.85\textwidth}{!}{%
\begin{tabular}{@{}l cc cc c@{}}
\multicolumn{6}{c}{\textit{Best Open-Source Model (Gemma-3 27B) Evaluation on 10\% Random Sample}} \\
\toprule
 & \multicolumn{2}{c}{\textbf{Idiom Usage Context}}
& \multicolumn{2}{c}{\textbf{Idiom Usage Type}}
& \\
\cmidrule(lr){2-3} \cmidrule(lr){4-5}
\textbf{Language}
& \textbf{Sent} & \textbf{Conv}
& \textbf{Fig} & \textbf{Lit}
& \textbf{Overall} \\
\midrule
\multicolumn{6}{@{}l}{\cellcolor{highres}\textbf{High-Resource Languages}} \\
\rowcolor{rowgray}
Japanese         &  57 & 84 & 91 & 45 & 69 \\
Russian          &  67 & 74 & 100 & 40 & 70 \\
\midrule
\multicolumn{6}{@{}l}{\cellcolor{medres}\textbf{Mid-Resource Languages}} \\
\rowcolor{rowgray}
Arabic (UAE)     & 70 & 88 & 90 & 67 & 78 \\
Indonesian       & 88 & 88 & 100 & 67 & 88 \\
\rowcolor{rowgray}
Vietnamese       & 100 & 100 & 100 & N/A & 100 \\
\midrule
\multicolumn{6}{@{}l}{\cellcolor{lowres}\textbf{Low-Resource Languages}} \\
\rowcolor{rowgray}
Arabic (Egypt)   &  68 & 82 & 100 & 47 & 75 \\
Arabic (Morocco)   &  50 & 82 & 84 & 44 & 65 \\
\rowcolor{rowgray}
Minangkabau      &  54 & 71 & 53 & 88 & 63 \\
Yoruba      &  17 & 21 & 16 & 29 & 19 \\
\midrule \rowcolor{rowgray}
\textbf{Average}
& \textbf{72} & \textbf{84}
& \textbf{90} & \textbf{59}
& \textbf{78} \\
\bottomrule
\noalign{\vskip 1pt}
\multicolumn{6}{c}{\small\textit{Sent = Sentence \quad\quad Conv = Conversation \quad\quad Fig = Figurative \quad\quad Lit = Literal}} \\
\end{tabular}%
}
\caption{Best overall open-source model (Gemma-3 27B) accuracy (\%) on a 10\% random sample of idioms from nine languages spanning all three resource tiers. Scores are reported across idiom usage contexts (Sentence, Conversation), idiom usage types (Figurative, Literal), and overall. Vietnamese is marked N/A for the literal column, as it contains no literal counterparts in \textbf{MIDI}.}
\label{tab:human_eval_open}
\end{table*}
\begin{table*}[t]
\centering
\small
\setlength{\tabcolsep}{6pt}
\renewcommand{\arraystretch}{1.2}
\resizebox{0.9\textwidth}{!}{%
\begin{tabular}{@{}l c c c cc c@{}}
\toprule
\textbf{Language}
& \textbf{Human}
& \textbf{Best Proprietary}
& \textbf{Best Open-Source}
& \multicolumn{2}{c}{\textbf{Gap (Human $-$ Best)}} \\
\cmidrule(lr){5-6}
&  &  &  & \textbf{Prop.} & \textbf{Open} \\
\midrule
\multicolumn{6}{@{}l}{\cellcolor{highres}\textbf{High-Resource Languages}} \\
\rowcolor{rowgray}
Japanese        &  88 & 71 \textit{(Gemini 2.5 Pro)} & 69 \textit{(Gemma-3)}      & +17 & +19 \\
Russian         &  99 & 65 \textit{(Gemini 2.5 Pro)} & 86 \textit{(Mixtral)}      & +34 & +13 \\
\midrule
\multicolumn{6}{@{}l}{\cellcolor{medres}\textbf{Mid-Resource Languages}} \\
\rowcolor{rowgray}
Arabic (UAE)    &  97 & 87 \textit{(GPT-5.2)}        & 81 \textit{(Llama-3.3)}    &  +10 & +16 \\
Indonesian      &  91 & 88 \textit{(Gemini 2.5 Pro)} & 88 \textit{(Gemma-3)}      &  +3 &  +3 \\
\rowcolor{rowgray}
Vietnamese      & 100 & 100 \textit{(Gemini 2.5 Pro \& GPT-5.2)} & 100 \textit{(Gemma-3)}      &  0 &  0 \\
\midrule
\multicolumn{6}{@{}l}{\cellcolor{lowres}\textbf{Low-Resource Languages}} \\
\rowcolor{rowgray}
Arabic (Egypt)  &  92 & 83 \textit{(Gemini 2.5 Pro)}        & 75 \textit{(Gemma-3 \& Llama-3.3)}      &  +9 & +17 \\
Arabic (Morocco)  &  89 & 62 \textit{(Gemini 2.5 Pro \& GPT-5.2)}        & 65 \textit{(Gemma-3)}      &  +27 & +24 \\
\rowcolor{rowgray}
Minangkabau     &  96 & 82 \textit{(Gemini 2.5 Pro)} & 63 \textit{(Gemma-3)}      & +14 & +33 \\
Yoruba     &  96 & 89 \textit{(Gemini 2.5 Pro)} & 23 \textit{(Mixtral)}      & +7 & +73 \\
\midrule \rowcolor{rowgray}
\textbf{Average}
& \textbf{94} & \textbf{81} & \textbf{72} & \textbf{+13} & \textbf{+22} \\
\bottomrule
\end{tabular}%
}
\caption{Comparison of overall human accuracy (\%) on a 10\% random sample of idioms against the best-performing proprietary and open-source models for each language. Model scores are representing performance on same samples; the top model per category is reported in parentheses. The rightmost columns show the accuracy gap (human minus model) in percentage points; positive values indicate human superiority. Humans outperform the best models on every language except Vietnamese, where Gemini 2.5 Pro \& Gemma-3 achieves parity, with the widest gaps concentrated in high- and low-resource tiers.}
\label{tab:human_vs_models}
\end{table*}

\begin{table*}[t]
\centering
\small
\setlength{\tabcolsep}{6pt}
\renewcommand{\arraystretch}{1.2}
\resizebox{0.9\textwidth}{!}{%
\begin{tabular}{@{}l c c c cc c@{}}
\toprule
\textbf{Language}
& \textbf{Human}
& \textbf{Best Proprietary}
& \textbf{Best Open-Source}
& \multicolumn{2}{c}{\textbf{Gap (Human $-$ Best)}} \\
\cmidrule(lr){5-6}
&  &  &  & \textbf{Prop.} & \textbf{Open} \\
\midrule
\multicolumn{6}{@{}l}{\cellcolor{highres}\textbf{High-Resource Languages}} \\
\rowcolor{rowgray}
Japanese        &  88 & 71 \textit{(Gemini 2.5 Pro)} & 70 \textit{(Gemma-3 \& Llama-3.3)}      & +17 ($\Delta$ 0) & +18 ($\Delta$ +1) \\
Russian         &  99 & 67 \textit{(Gemini 2.5 Pro)} & 80 \textit{(Mixtral)}      & +32 ($\Delta$ +2) & +19 ($\Delta$ $-$6) \\
\midrule
\multicolumn{6}{@{}l}{\cellcolor{medres}\textbf{Mid-Resource Languages}} \\
\rowcolor{rowgray}
Arabic (UAE)    &  97 & 87 \textit{(GPT-5.2)}        & 82 \textit{(Llama-3.3)}    &  +10 ($\Delta$ 0) & +15 ($\Delta$ +1) \\
Indonesian      &  91 & 91 \textit{(Gemini 2.5 Pro)} & 89 \textit{(Gemma-3)}      &  0 ($\Delta$ +3) &  +2 ($\Delta$ +1) \\
\rowcolor{rowgray}
Vietnamese      & 100 & 99 \textit{(Gemini 2.5 Pro)} & 99 \textit{(Gemma-3)}      &  +1 ($\Delta$ $-$1) &  +1 ($\Delta$ $-$1) \\
\midrule
\multicolumn{6}{@{}l}{\cellcolor{lowres}\textbf{Low-Resource Languages}} \\
\rowcolor{rowgray}
Arabic (Egypt)  &  92 & 85 \textit{(GPT-5.2)}        & 80 \textit{(Gemma-3)}      &  +7 ($\Delta$ +2) & +12 ($\Delta$ +5) \\
Arabic (Morocco)  &  89 & 70 \textit{(Gemini 2.5 Pro \& GPT-5.2)}        & 67 \textit{(Gemma-3 \& Llama-3.3)}      &  +19 ($\Delta$ +8) & +22 ($\Delta$ +2) \\
\rowcolor{rowgray}
Minangkabau     &  96 & 86 \textit{(Gemini 2.5 Pro)} & 49 \textit{(Gemma-3)}      & +10 ($\Delta$ +4) & +47 ($\Delta$ $-$14) \\
Yoruba     &  96 & 80 \textit{(Gemini 2.5 Pro)} & 25 \textit{(DeepSeek-R1)}      & +16 ($\Delta$ $-$9) & +71 ($\Delta$ +2) \\
\midrule \rowcolor{rowgray}
\textbf{Average}
& \textbf{94} & \textbf{82} ($\Delta$ $-$1) & \textbf{71} ($\Delta$ +1) & \textbf{+12} ($\Delta$ +1) & \textbf{+23} ($\Delta$ $-$1) \\
\bottomrule
\end{tabular}%
}
\caption{Comparison of human accuracy (\%) on a 10\% random sample of idioms against the best-performing proprietary and open-source models evaluated on the \emph{full} per-language dataset (drawn from Table~\ref{tab:per_language_accuracy}). The rightmost columns show the accuracy gap (human minus model) in percentage points; positive values indicate human superiority. The \textbf{$\Delta$} value in parentheses next to each gap denotes the difference between this gap and the corresponding gap in Table~\ref{tab:human_vs_models}, which compares the same human scores against model performance on the matched 10\% sample.}
\label{tab:human_vs_models_full}
\end{table*}

\section{Dataset Details}
\label{appx:dataset_extra}

This appendix provides additional details about the composition of \textbf{MIDI}. Table~\ref{tab:midi_examples} presents example instances from the dataset. Table~\ref{tab:dataset_stats} reports per-language dataset statistics, including idiom counts, literal-counterpart coverage, and the number of figurative and literal contexts across sentence and dialogue formats. In addition, Table~\ref{tab:idiom_length_stats} reports descriptive statistics for both idiomatic and literal expressions in sentence- and dialogue-level contexts.

\begin{table*}[ht!]
    \centering
    \footnotesize
    \renewcommand{\arraystretch}{1.3}
    \setlength{\tabcolsep}{4pt}.
    \resizebox{\textwidth}{!}{%
    \begin{tabular}{@{}
        >{\raggedright\arraybackslash}p{0.07\textwidth}
        >{\raggedright\arraybackslash}p{0.10\textwidth}
        >{\raggedright\arraybackslash}p{0.10\textwidth}
        >{\raggedright\arraybackslash}p{0.07\textwidth}
        >{\raggedright\arraybackslash}p{0.10\textwidth}
        >{\raggedright\arraybackslash}p{0.17\textwidth}
        >{\raggedright\arraybackslash}p{0.10\textwidth}
        >{\raggedright\arraybackslash}p{0.19\textwidth}
        >{\centering\arraybackslash}p{0.03\textwidth}
    @{}}
    \toprule
    \textbf{Idiom} & \textbf{Definition (EN)} & \textbf{Definition (Native)} & \textbf{Type} & \textbf{Sentence} & \textbf{Dialogue} & \textbf{Question} & \textbf{Choices} & \textbf{Ans.} \\
    \midrule
\viet{Bắt cá hai tay} & Cunning, greedy action, want to have many things at one time & \viet{Hành động khôn lỏi, tham lam, muốn có nhiều thứ} & figurative & \viet{Anh ấy bị phát hiện bắt cá hai tay nên cả hai cô gái đều chia tay anh ta.} & 
A: \viet{Nghe nói Minh cùng lúc tán tỉnh hai người bạn cùng lớp?} \newline B: \viet{Đúng rồi, cậu ấy không quyết định nổi ai cả.} \newline A: \viet{Rõ là bắt cá hai tay.} & What does the phrase \textit{\viet{Bắt cá hai tay}} mean? & A. \viet{Tài năng bắt được cá bằng cả hai tay} \newline B. \viet{Chỉ người chăm chỉ làm nhiều việc cùng lúc} \newline C. \viet{Bắt được hai con cá cùng một lúc} \newline D. \viet{Hành động khôn lỏi, muốn có nhiều thứ} & D \\
    \addlinespace[4pt]
    \hdashline
    \addlinespace[4pt]
ẹja n bakan? & is it positive or negative & so bọsi abi ko bọsi & literal & ṣe ẹja n bakan lo fe lori ounje re? & A : baba lagabja, booni elo ounje aleyi, ẹja abi akan? baba lagabja: ah eja ni o, ti mo ba ri ponmo na iyen na a lọ \newline A: oda, olohun a ṣe iyanu & What does the phrase \textit{ẹja n bakan?} mean? & A. se eja tabi akan ni won o fi jehun \newline B. se baba lagbaja fẹran akan tabi ẹja? \newline C. on bere pe se o lo eja tabi akan \newline D. se oro ti won so bọsi tabi ko bọsi & A \\
\bottomrule
\end{tabular}%
}
    \caption{Example instances from the MIDI dataset.}
    \label{tab:midi_examples}
\end{table*}

\begin{table*}[t]
\centering
\renewcommand{\arraystretch}{1.2}
\resizebox{\textwidth}{!}{%
\begin{tabular}{@{}l cc c cc cc c@{}}
\toprule
\textbf{Language}
& \multicolumn{2}{c}{\textbf{Idiom Inventory}}
& \textbf{Literal Cov.\ (\%)}
& \multicolumn{2}{c}{\textbf{Figurative Contexts}}
& \multicolumn{2}{c}{\textbf{Literal Contexts}}
& \textbf{Total Contexts} \\
\cmidrule(lr){2-3} \cmidrule(lr){5-6} \cmidrule(lr){7-8}
& \textbf{\#Idioms} & \textbf{\#Lit.\ Counterparts}
&
& \textbf{Sentence} & \textbf{Dialogue}
& \textbf{Sentence} & \textbf{Dialogue}
& \\
\midrule
\multicolumn{9}{@{}l}{\cellcolor{highres}\textbf{High-Resource Languages}} \\
\rowcolor{rowgray}
Chinese          & 300 & 300 & 100.0 & 300 & 300 & 300 & 300 & 1{,}200 \\
Japanese         & 115 & 109 &  94.8 & 115 & 115 & 109 & 109 &   448 \\
\rowcolor{rowgray}
Russian          & 213 & 208 &  97.7 & 213 & 213 & 208 & 208 &   842 \\
\midrule
\multicolumn{9}{@{}l}{\cellcolor{medres}\textbf{Mid-Resource Languages}} \\
\rowcolor{rowgray}
Arabic (UAE)     & 100 &  98 &  98.0 & 100 & 100 &  98 &  98 &   396 \\
Indonesian       & 108 &  57 &  52.7 & 108 & 108 &  57 &  57 &   330 \\
\rowcolor{rowgray}
Vietnamese       & 100 &   0 &   0.0 & 100 & 100 &   0 &   0 &   200 \\
\midrule
\multicolumn{9}{@{}l}{\cellcolor{lowres}\textbf{Low-Resource Languages}} \\
\rowcolor{rowgray}
Arabic (Egypt)   & 100 &  84 &  84.0 & 100 & 100 &  84 &  84 &   368 \\
Arabic (Morocco) &  99 &  93 &  93.9 &  99 &  99 &  93 &  93 &   384 \\
\rowcolor{rowgray}
Arabic (Syria)   & 100 & 100 & 100.0 & 100 & 100 & 100 & 100 &   400 \\
Persian          & 102 &  82 &  80.4 & 102 & 102 &  82 &  82 &   368 \\
\rowcolor{rowgray}
Javanese         & 104 &  96 &  92.3 & 104 & 104 &  96 &  96 &   400 \\
Kannada          & 198 & 198 & 100.0 & 198 & 198 & 198 & 198 &   792 \\
\rowcolor{rowgray}
Kazakh           & 100 & 100 & 100.0 & 100 & 100 & 100 & 100 &   400 \\
Minangkabau      & 100 &  49 &  49.0 & 100 & 100 &  49 &  49 &   298 \\
\rowcolor{rowgray}
Sundanese        & 100 &  72 &  72.0 & 100 & 100 &  72 &  72 &   344 \\
Tamil            &  99 &  95 &  96.0 &  99 &  99 &  95 &  95 &   388 \\
\rowcolor{rowgray}
Telugu           & 139 & 137 &  98.6 & 139 & 139 & 137 & 137 &   552 \\
Yoruba           & 101 &  33 &  32.7 & 101 & 101 &  33 &  33 &   268 \\
\midrule \rowcolor{rowgray}
\textbf{Total}
& \textbf{2{,}278} & \textbf{1{,}911} & \textbf{82.3}
& \textbf{2{,}278} & \textbf{2{,}278}
& \textbf{1{,}911} & \textbf{1{,}911}
& \textbf{8{,}378} \\
\bottomrule
\end{tabular}%
}
\caption{Per-language dataset statistics. For each idiom, a figurative sentence and dialogue are provided; for idioms with a literal counterpart, an additional literal sentence and dialogue are provided. \textbf{Literal Cov.} denotes the percentage of idioms with a literal counterpart. Rows are grouped by language resource availability.}
\label{tab:dataset_stats}
\end{table*}
\definecolor{highres}{HTML}{C8E6C9}    
\definecolor{medres}{HTML}{FFF9C4}     
\definecolor{lowres}{HTML}{FFCDD2}     
\definecolor{rowgray}{HTML}{F5F5F5}    

\begin{table*}[t]
\centering
\small
\setlength{\tabcolsep}{4pt}
\renewcommand{\arraystretch}{1.15}
\resizebox{1\textwidth}{!}{%
\begin{tabular}{@{}l cc cc cc cc@{}}
\toprule
\textbf{Language} 
& \multicolumn{2}{c}{\textbf{Figurative (Sentence)}} 
& \multicolumn{2}{c}{\textbf{Figurative (Dialogue)}} 
& \multicolumn{2}{c}{\textbf{Literal (Sentence)}} 
& \multicolumn{2}{c}{\textbf{Literal (Dialogue)}} \\
\cmidrule(lr){2-3} \cmidrule(lr){4-5} \cmidrule(lr){6-7} \cmidrule(lr){8-9}
& \textbf{\#Wrds} & \textbf{\#Chrs}
& \textbf{\#Wrds} & \textbf{\#Chrs}
& \textbf{\#Wrds} & \textbf{\#Chrs}
& \textbf{\#Wrds} & \textbf{\#Chrs} \\
\midrule

\multicolumn{9}{@{}l}{\cellcolor{highres}\textbf{High-Resource Languages}} \\
\rowcolor{rowgray}
Chinese        & 17.12 & 25.67 & 40.83 & 61.19 & 32.99 & 47.54 & 49.07 & 72.62 \\
Japanese       & 13.00 & 19.37 & 37.24 & 59.54 & 15.29 & 23.08 & 48.43 & 75.60 \\
\rowcolor{rowgray}
Russian        & 12.05 & 73.77 & 31.46 & 187.02 & 13.05 & 82.49 & 33.49 & 200.80 \\

\midrule
\multicolumn{9}{@{}l}{\cellcolor{medres}\textbf{Mid-Resource Languages}} \\
\rowcolor{rowgray}
Arabic (UAE)   & 11.23 & 60.63 & 22.36 & 120.40 & 11.59 & 63.52 & 25.81 & 139.37 \\
Indonesian     & 11.38 & 77.56 & 24.81 & 146.81 & 10.86 & 74.28 & 27.81 & 161.89 \\
\rowcolor{rowgray}
Vietnamese     & 19.72 & 85.86 & 36.63 & 159.38 & 11.00 & 58.00 & 7.00  & 41.00 \\

\midrule
\multicolumn{9}{@{}l}{\cellcolor{lowres}\textbf{Low-Resource Languages}} \\
\rowcolor{rowgray}
Arabic (Egypt)   & 8.78 & 43.15 & 22.35 & 117.36 & 7.44 & 37.07 & 25.88 & 137.64 \\
Arabic (Morocco) & 11.92 & 61.86 & 23.49 & 128.55 & 10.58 & 56.71 & 28.47 & 160.89 \\
\rowcolor{rowgray}
Arabic (Syria)   & 7.55 & 38.53 & 22.32 & 118.00 & 8.84 & 46.86 & 25.81 & 136.32 \\
Persian          & 12.99 & 63.64 & 26.33 & 134.36 & 12.76 & 61.12 & 31.84 & 160.94 \\
\rowcolor{rowgray}
Javanese         & 13.45 & 85.06 & 28.22 & 162.50 & 10.80 & 67.71 & 29.20 & 168.42 \\
Kannada          & 6.74 & 54.13 & 30.55 & 235.51 & 6.18 & 48.06 & 22.82 & 174.18 \\
\rowcolor{rowgray}
Kazakh           & 7.28 & 49.64 & 27.47 & 187.98 & 6.35 & 42.17 & 24.09 & 162.36 \\
Minangkabau      & 8.42 & 50.23 & 30.19 & 159.67 & 9.47 & 57.24 & 29.61 & 161.41 \\
\rowcolor{rowgray}
Sundanese        & 13.53 & 79.28 & 25.48 & 150.36 & 13.12 & 77.53 & 28.11 & 163.88 \\
Tamil            & 7.71 & 64.05 & 20.92 & 170.01 & 5.93 & 48.31 & 21.62 & 177.80 \\
\rowcolor{rowgray}
Telugu           & 7.16 & 53.66 & 15.37 & 93.73  & 6.15 & 46.58 & 19.68 & 113.92 \\
Yoruba           & 12.50 & 52.35 & 26.77 & 115.20 & 11.20 & 57.87 & 26.61 & 115.24 \\

\midrule \rowcolor{rowgray}
\textbf{Average} 
& \textbf{11.25} & \textbf{57.69}
& \textbf{27.38} & \textbf{139.31}
& \textbf{11.31} & \textbf{55.34}
& \textbf{28.08} & \textbf{140.24} \\

\bottomrule
\end{tabular}%
}
\caption{Average length statistics (words and characters) for idiom usage across sentence and dialogue contexts, separated by figurative and literal interpretations and grouped by language resource level.}
\label{tab:idiom_length_stats}
\end{table*}

\end{document}